\documentclass[fleqn,10pt]{wlscirep}
\pdfoutput=1

\usepackage[utf8]{inputenc}
\usepackage[T1]{fontenc}
\usepackage{lineno}
\usepackage{subfiles}
\usepackage{multirow}
\usepackage{graphicx}
\usepackage{placeins}
\usepackage{listings}
\usepackage{hyperref}
\newcommand{\subjective}{\textsc{\textsc{subjective}}}
\newcommand{\assessment}{\textsc{\textsc{assessment\_and\_plan}}}
\newcommand{\exam}{\textsc{\textsc{objective\_exam}}}
\newcommand{\results}{\textsc{\textsc{objective\_results}}}

\newcommand{\acidemo}{\textsc{\textsc{aci-bench}}}
\newcommand{\consult}{\textsc{\textsc{consult}}}

\newcommand{\virtassist}{\textbf{virtassist}}
\newcommand{\virtscribe}{\textbf{virtscribe}}
\newcommand{\aci}{\textbf{aci}}

\newcommand{\medcon}{\textbf{\textsc{\textsc{medcon}}}}

\title{\acidemo{}: a Novel Ambient Clinical Intelligence Dataset for Benchmarking Automatic Visit Note Generation}

\author[1,*]{Wen-wai Yim}
\author[2]{Yujuan Fu}
\author[1]{Asma Ben Abacha}
\author[3]{Neal Snider}
\author[1]{Thomas Lin}
\author[2]{Meliha Yetisgen}
\affil[1]{Microsoft, Health and Life Sciences AI, Redmond, 98052, USA}
\affil[2]{University of Washington, Biomedical and Health Informatics, Seattle, 98109, USA}
\affil[3]{Nuance Communications, Healthcare R\&D, Burlington, 01803, USA}

\affil[*]{corresponding author(s): Wen-wai Yim (yimwenwai@microsoft.com)}

\begin{abstract}
Recent immense breakthroughs in generative models such as in  GPT4 have precipitated re-imagined ubiquitous usage of these models in all applications. One area that can benefit by improvements in artificial intelligence (AI) is healthcare. The note generation task from doctor-patient encounters, and its associated electronic medical record documentation, is one of the most arduous time-consuming tasks for physicians. It is also a natural prime potential beneficiary to advances in generative models. However with such advances,  benchmarking is more critical than ever. Whether studying model weaknesses or developing new evaluation metrics, shared open datasets are an imperative part of understanding the current state-of-the-art. Unfortunately as clinic encounter conversations are not routinely recorded and are difficult to ethically share due to patient confidentiality, there are no sufficiently large clinic dialogue-note datasets to benchmark this task. Here we present the Ambient Clinical Intelligence Benchmark (\acidemo{}) corpus, the largest dataset to date tackling the problem of AI-assisted note generation from visit dialogue. We also present the benchmark performances of several common state-of-the-art approaches.
\end{abstract} 

\begin{document}

\flushbottom
\maketitle
%  Click the title above to edit the author information and abstract

\thispagestyle{empty}

% \noindent Please note: Abbreviations should be introduced at the first mention in the main text – no abbreviations lists or tables should be included. Structure of the main text is provided below.

\section{Background \& Summary}

Healthcare needs are an inescapable facet of daily life. Current patient care at the medical facilities requires involvement not only from a primary care provider, but also from pharmacy, billing, imagining, labs, and specialist care. For every encounter, a clinical note is created as documentation of clinician-patient discussions, patient medical conditions. They serve as a vital record for clinical care and communication with patients and other members of the care team, as well as outline future plans, tests, and treatments. Similar to typical meeting summaries, these documents should highlight important points while compressing itemized instances into condensed themes; unlike typical meeting summaries, clinical notes are purposely and technically structured into semi-structured documents, contain telegraphic and bullet-point phrases, use medical jargon that do not appear in the original conversation, and will reference outside information often from the electronic medical record, including prose-written content or injections of structured data.

While the widespread adoption of electronic health records (EHR's), spurred by the HITECH Act of 2009, has led to greater health information availability and interoperability, it has also spawned a massive documentation burden shifted to clinicians. Physicians have expressed concerns that writing notes in electronic health records (EHRs) takes more time than using traditional paper or dictation methods. As a result, notes may not be completed and accessible to other team members until long after rounds \cite{mcdonald_use_2014}. Furthermore, as another unintended consequence of EHR use complications, electronic notes have been criticized for their poor readability, completeness, and excessive use of copy and paste \cite{embi_computerized_2013}. To save time and adequately capture details, clinicians may choose to write their notes during their time with a patient. This may detract from the clinicians' attention toward the patient (e.g. in reading non-verbal cues), and may leave patients feeling a want of empathy \cite{toll_cost_2012}. Alternatively, some clinicians or provider systems may hire medical assistants or scribes to partake in some or all of the note creation process, which has been linked with improved productivity, increased revenue, and improved patient-clinician interactions\cite{shultz_use_2015}. However such systems are both costly and, more importantly, often require a substantial investment in time from the providers in managing and training their scribes \cite{tran_how_2020} -- a problem that is often multiplied by the high attrition rates in the field.

One promising solution is the use of automatic summarization to capture and draft notes, before being reviewed by a clinician. This technology has attracted increasing attention in the last 5 years as a result of several key factors: (1) the improvement of speech-to-text technology, (2) widespread adoption of electronic medical records in the United States, (3) the rise of transformer models. Several works have adopted early technology in this area, including use of statistical machine translation methods, use of RNNs, transformers, and pre-trained transformer models\cite{finley-etal-2018-dictations,enarvi-etal-2020-generating,krishna-etal-2021-generating,zhang-etal-2021-leveraging-pretrained,michalopoulos-etal-2022-medicalsum,yim-yetisgen-2021-towards}.

However, a massive bottleneck in understanding the state-of-the-art is the lack of publicly share-able data to train and evaluate\cite{quiroz_challenges_2019}. This challenge is inherent in the required data's characteristics as (1) meeting audio and transcripts from medical encounters are not typically recorded and saved, and (2) medical information is highly personal and sensitive data and cannot be easily, ethically shared publicly. Private companies may construct or acquire their own private datasets; however, results and algorithms cannot be systematically compared. Recent ground-breaking performances by large language models such as ChatGPT and GPT4 provide promising general model solutions; however without common datasets that may be studied publicly it would be impossible for the scientific community to understand strength, weaknesses, and future directions.

\begin{table}[ht]
\centering
\begin{tabular}{l|p{45mm}|l|l|l|l}
\hline
dataset & description & src-len (tok/turns) & target-len (tok/sent) & size & open\\
\hline
MTS-dialogue \cite{benabacha2023empirical} & dialogue-note snippets where conversations are created using clinical note sections & 142/9 & 48/3 & 1701 & Y\\
primock57 \cite{papadopoulos-korfiatis-etal-2022-primock57} & role-played dialogue-note pairs & 1489/97 & 161/23 & 57 & Y\\
\textbf{\acidemo{}[this work]} & role-played dialogue-note pairs & 1302/55 & 490/49 & 207 & Y\\
\hline
3M Health\cite{zhang-etal-2021-leveraging-pretrained} & dialogue-note pairs where notes are created using conversations & -/- & -/- (hpi only) & 1342 & N \\
Abridge \cite{krishna-etal-2021-generating}  & dialogue-note pairs where notes are created using conversations & 1500/- & -/27 & 6862 & N \\
Augmedix \cite{yim-yetisgen-2021-towards} & real clinical dialogue-note pairs & -/175 & -/47 & 500 & N \\
emr.ai \cite{finley-etal-2018-dictations} & real clinical dictation-note pairs & 616/1 & 550/- & 9875 & N \\
Nuance \cite{enarvi-etal-2020-generating} & real clinical dialogue-note pairs & 972 avg/- & 452 total/-\footnote{the authors model sections of the note differently. The number of sources and note sections are different. Here we approximate the average note length by adding the average section lengths together. Average source length was approximated by averaging the sources for different sections.} & 802k & N \\
\hline
\end{tabular}
\caption{\label{tab:comparable-corpora}Comparable corpora for doctor-patient dialogue2note generation. The majority of datasets are proprietary and unshare-able for community evaluation. (src-len=source/transcript length, target-len=target/note length, -=unreported)}
\end{table}

In this paper, we present the Ambient Clinical Intelligence Benchmark (\acidemo{}) corpus. The corpus, created from domain experts, is designed to model three variations of model-assisted clinical note generation from doctor-patient conversations. These include conversations with (a) calls to a virtual assistant (e.g. required use of wake words or prefabricated, canned phrases), (b) unconstrained directions or discussions with a scribe, and (c) natural conversations  between a doctor and patient. We also provide data to experiment between using human transcription and automatic speech recognition (ASR); or between ASR and corrected ASR. Table 1 shows a comparison of the 8 corpora described in state-of-the-art work. Only two other similar corpora are publicly available. primock57\cite{papadopoulos-korfiatis-etal-2022-primock57} contains a small set of 57 encounters. MTS-dialog\cite{benabacha2023empirical} contains $\sim$1700 samples however its focus is on on dialogue snippets rather than full encounters. To our knowledge, \acidemo{} is the largest and most comprehensive corpus publicly available for model-assisted clinical note generation. 

In the following sections, we provide details of the \acidemo{} Corpus. We (1) discuss the dataset construction and cleaning, (2) provide statistics and the corpus structure, (3) describe our content validation methods and comparison with real data, (4) quantify several diverse baseline summarization methods on this corpus.

\section{Methods}
\label{sec:methods}

\subsection{Data Creation}

Clinical notes may be written by the physician themselves or in conjunction with a medical scribe or assistant; alternatively physicians may choose to dictate the contents of an entire note to a human transcriptionists or an automatic dictation tool. In cases with human intervention, scribe-assisted or transcriptionist-assisted cases, physician speech may include a mixture of commands (e.g. ``newline'', ``add my acne template''), free-text requiring almost word-for-word copying (e.g. ``To date, the examinee is a 39 year-old golf course maintenance worker'') \cite{finley-etal-2018-dictations}, or free-text communication to the medical assistance (e.g. ``let's use my normal template, but only keep the abnormal parts'', ``can you check the date and add that in?''). With trained medical scribes participating in the clinic visit, in addition to directions from the doctor, they are expected to listen in on the patient-doctor dialogue and generate clinical note text independently.
To mirror this reality, the \acidemo{} corpus consists of three subsets representing common modes of  note generation from doctor-patient conversations:\\

\textbf{virtual assistant (virtassist)}: In this mode, the doctor may use explicit terms to activate a virtual assistance device (e.g. ``Hey Dragon show me the diabetes labs'') during the visit. This necessitates some behavioral changes on the part of the provider. \\

\textbf{virtual scribe (virtscribe)}: In this mode, the doctor may  expect a separate scribe entity (automated or otherwise) to help create the clinical note. This subset is characterized by pre-ambles (e.g. short patient descriptions prior to a visit) and after-visit dictations (e.g. used to specify non-verbal parts of the visit such as the physical exam or to dictate the assessment and plan). The rest of the doctor-patient conversation will be natural and undisturbed.\\

\textbf{ambient clinical intelligence (aci)}: This data is characterized by natural conversation between a patient and a doctor; without explicit calls to a virtual assistant or additional language addressed to a scribe.\\

Transcripts from subsets \virtassist{} and \virtscribe{} were created by a team of 5+ medical experts including medical doctors, physician assistance, medical scribes, and clinical informaticians based on experience and studying real encounters. Subset \aci{} was created with a certified doctor and a volunteer lay person, who must role-play a real doctor-patient encounter, given a list of symptom prompts. Clinical notes were generated using an automatic note generation system and checked and re-written by domain experts (e.g. medical scribes, or physicians). The \virtscribe{} dataset includes the human transcription as well as an ASR transcript; meanwhile the \virtassist{} and \aci{} subsets were created with only a human transcription and ASR transcript available, respectively.

\subsubsection{Data Cleaning and Annotation}

Our final dataset was distilled from encounters originally created for marketing demonstration purposes. During this initial dataset creation, imaginary EHR injections were placed within the note to contribute to realism, though many without  basis from the conversation. Although EHR inputs, independent from data intake from a conversation, are a critical aspect of real clinical notes, in this dataset we do not model EHR input or output linkages with the clinical note (e.g. smart links to structured data such as vitals values, structured survey data, order codes, and diagnosis codes).

\begin{table}[h]
\centering
\begin{tabular}{l|l|p{70mm}}
\hline
type & annotated & example \\
\hline
 dates (none mentioned) & Y & PSA 0.6 ng/mL[, 05/25/2021]\\
exam & Y & [Consitutional: Well-developed, well-nourished, in no apparent distress]\\

 & Y & Neck: [Supple without thyromegaly or lymphadenopathy.] No carotid bruits appreciable.\\
medication context & Y & 1 tablet [by oral route] daily\\
 medical reasoning & Y & recommended that we obtain an MRI of the right shoulder [to evaluate for a possible rotator cuff tear].\\
& Y & referred her to formal physical therpay [to strengthen her right shoulder]\\
 patient acquiescence & Y & [All questions were answered.]\\
 & Y & [The patient understands and agrees with the recommended medical treatment plan]\\
 review of system & Y & Ears, Nose, Mouth and Throat: [Denies ear pain, hearing loss, or discharge.] Endorses nasal congestion from allergies.\\
 vitals & Y & [Blood Pressure:124/82 mmHg]\\
\hline
dates (year kept if only month mentioned) & N & 03/[2022] \\
higher granularity problem/test/treatment & N & diabetes [type II] \\
 & N & [3 views] of the shoulder \\
 & N & MRI [of the head] \\
measurements & N & 25 [mg/dl] \\
patient name/age & N & [John Smith] is a [53]-year-old male \\
other names & N & he was seen by Jane [Smith, PA-C] \\
\hline
\end{tabular}
\caption{\label{tab:unsupported-text}Examples of unsupported text (demarked by square brackets). In the original demo data, these items were added for realism without basis in the source text, the doctor-patient conversation. Some unsupported items were purposely left unmarked in cases where removal would lead to note quality / meaning degradation. After human annotated text-span level identification, these were automatically removed from the clinical note.}
\end{table}

In order to identify unsupported information of note text to the transcript, we created systematic annotation guidelines for labeling unsupported note sentences. These unsupported information included items such as reasoning for treatment (which may not be part of the original conversation) or could be information from imaginary EHR inputs (e.g. vitals). Examples of the different types of unsupported information are included in Table \ref{tab:unsupported-text}. We tasked four independent annotators with medical backgrounds to complete this task. The partial span overlap agreement was 0.85 F1. Marked text spans were removed during automatic processing.\\
\indent Because the datasets were originally created and demonstrated for a short period, as such, these notes were created under greater time constraints and less review. To ensure quality, four annotators identified and corrected note errors, such as inconsistent values. Finally, as the \acidemo{} dataset used ASR transcripts, there were cases where the note and the transcript information would conflict due to ASR errors. For example, ``hydronephosis'' in the clinical note may be wrongly automatically transcribed as ``high flow nephrosis''. Another example may be a names; ``Castillo'' may be transcribed as ``kastio''. As part of this annotation, we tasked annotators to identify these items and provide corrections. After annotation, the data was processed such that note errors were corrected and unsupported note sentences were removed. To study the effect of ASR errors, ASR transcripts were processed into two versions: (a) original and (b) ASR-corrected (ASR outputs corrected by humans). After automatic processing, encounters were again manually reviewed for additional misspelling and formatting issues.

\subsection{Note Division Definition}

Motivated by a need to simplify clinical note structure, improve sparsity problems, and simplify evaluation, in this section, we describe our system for segmenting a full clinical note into continuous divisions.

\begin{figure}[bt]
\centering
\includegraphics[keepaspectratio=true,scale=0.3]{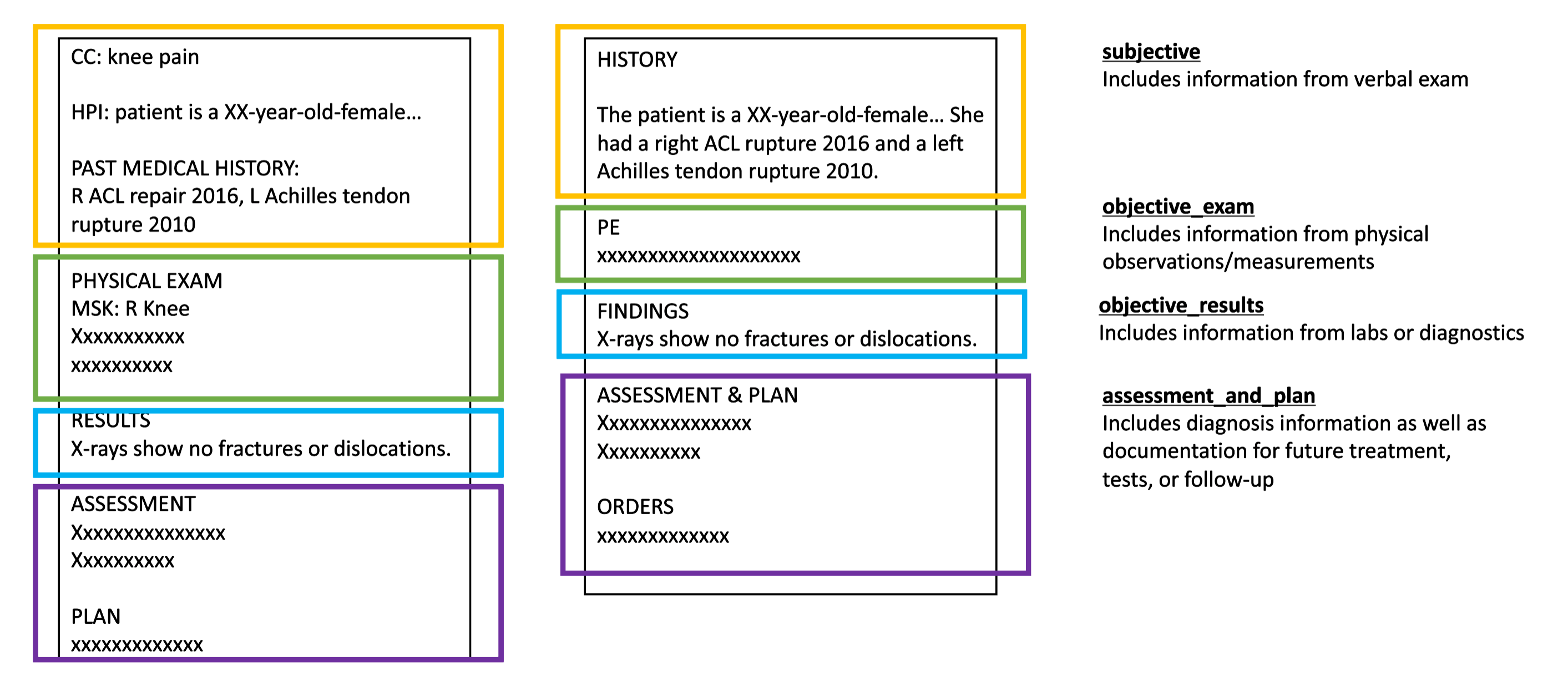}
\caption{Note division example. The same content in a clinical note can appear under different sections. As an example, in the left note, ``past medical history'' contents are written in the ``history'' portion of the note on the right. To seperate the full note target into smaller text and minimize data sparsity problems if modeling by individual sections, notes are partitioned into separate \subjective{}, \exam{}, \results{}, and \assessment{} continuous divisions. This also allows evaluation and generation at a higher granularity compared to a full note level.}\label{subnote-eval}
\end{figure}

\begin{figure}[h]
\centering
\includegraphics[keepaspectratio=true,scale=0.3]{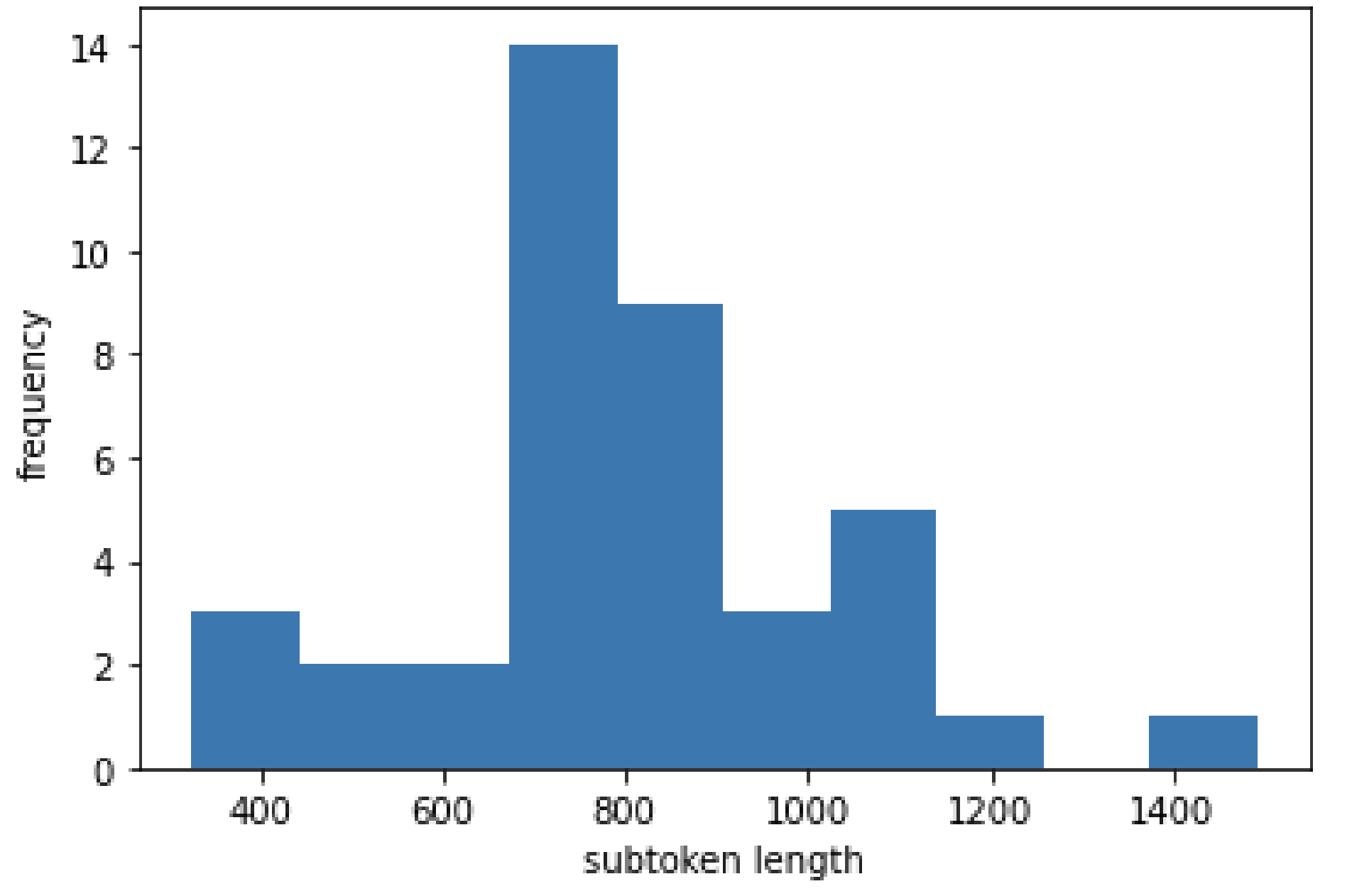}
\caption{BERT subtoken lengths of concatenated gold/system summaries (test1 Text-davinci-003 system) for doctor-patient dialogue to clinical note generation task. As embedding-based models require encoding the concatenated reference and hypothesis, on this dataset it would be difficult to fairly evaluate the corpus using current pretrained BERT models which have a 512 subtoken limit. }\label{subtok-hist}
\end{figure}

Clinical notes are semi-structural documents with hierarchical organization. Each physician, department, and institution may have their own set of commonly used formats. However, no universal standard exists\cite{denny_development_2008}. The same content can appear in multiple forms structured under different formats. This is illustrated in the subjective portions of two side-by-side notes in Figure \ref{subnote-eval}. In this example, contextual medical history appears in their own sections (e.g. ``current complaint (cc)'' and ``history of present illness (hpi)'', ''past medical history'') in the report on the left; and merged into one history section in the report on the right.  These variations in structure pose challenges for both generation and evaluation. Specifically, if evaluating by fine-grained sections in the reference, it is possible that generated notes may include the same content in other sections. Likewise, generating with fine-grained sections would require sufficient samples from each section; however as not every note has every type of section -- the sample size becomes sparser. Finally, it is important to note, current state-of-the-art pre-trained embedding based evaluation metrics (e.g. bertscore, bleurt, bart-score) are limited by the original trained sequence length which are typically shorter than our full document lengths. This is illustrated in Figure \ref{subtok-hist}, where for one system (Text-davinci-003) the length of the concatenated reference and system summaries will typically far exceed the typical pre-trained BERT-based 512 subtoken limit.

To simplify training and evaluation, as well as maintain larger samples of data, we partition notes and group multiple sections together into four divisions, as shown in Figure \ref{subnote-eval} These divisions were inspired by the SOAP standard, where the \subjective{} includes items taken during verbal exam and typically written in the chief complaint, history of present illness, and past social history; the \exam{} includes content from the physical examination on the day of the visit; the \results{} includes diagnostics taken prior to the visit, including laboratory or imaging results; and the \assessment{} includes the doctor's diagnosis and planned tests and treatments \cite{podder_soap_2023}. In our dataset, the divisions are contiguous and appear in the order previously introduced. Another practical benefit of partitioning the note into contiguous divisions is the greater ability to leverage pretrained sequence-to-sequence models, typically trained with shorter sequences. Furthermore, evaluation at a sub-note level allows a greater resolution for assessing performances.

\subsection{Data Statistics}
The full dataset was split into train, validation, and three test sets. Each subset was represented in the splits through randomized stratefied sampling. The test sets 1 and 2 corresponds to the test sets from  ACL ClinicalNLP MEDIQA-Chat 2023\footnote{\url{https://github.com/abachaa/MEDIQA-Chat-2023}} TaskB and TaskC, respectively. Test 3 corresponds to TaskC of CLEF MEDIQA-SUM 2023\footnote{\url{https://www.imageclef.org/2023/medical/mediqa}}. The frequency of each data split are shown in Table \ref{tab:corpus-statistics}.

\begin{table}[ht]
\centering
\begin{tabular}{l l l l l l}
\hline
 & train & valid & test1 & test2 & test3\\
\hline
number encounters & 67 & 20 & 40 & 40 & 40 \\
\hline
 \underline{dialogue} &  &  &  &  & \\
avg number turns & 56 & 53 & 52 & 56 & 58 \\
avg length (tok) & 1301 & 1221 & 1231 & 1382 & 1334 \\
\hline
 \underline{note} &  &  &  &  & \\
avg length (tok) & 483 & 492 & 476 & 500 & 505 \\
avg length (sentences) & 48 & 49 & 47 & 50 & 50 \\
\# subjective & 67 & 20 & 40 & 40 & 40 \\
\# objective\_exam & 64 & 19 & 40 & 39 & 39 \\
\# objective\_results & 53 & 18 & 32 & 29 & 27 \\
\# asessment\_and\_plan & 67 & 20 & 40 & 40 & 40 \\
\hline
 \underline{subset} &  &  &  &  & \\
\# \virtassist{} & 20 & 5 & 10 & 10 & 10 \\
\# \virtscribe{} & 12 & 4 & 8 & 8 & 8\\
\# \aci{} & 35 & 11 & 22 & 22 & 22\\
\hline
\end{tabular}
\caption{\label{tab:corpus-statistics} Corpus statistics}
\end{table}

\FloatBarrier

\section{Data Records}

The \acidemo{} Corpus can be found at [LINK TO BE UPDATED]. Code for pre-processing, evaluation, and running baselines can be found in [LINK TO BE UPDATED].

\subsection{Folder and naming organization}
Data used in the ACL-clinicalnlp MEDIQA-CHAT and CLEF  MEDIQASUM challenges are located in the \textit{challenge\_data} folder, whereas ASR experiment data is located in the \textit{src\_experiment\_data} folder. Each data split has  two associated files: a metadata and a data file (further described below). Train, validation, test1, test2, and test3 data files are prefixed with the following names: train, valid, clinicalnlp\_taskB\_test1, clinicalnlp\_taskC\_test2, and clef\_taskC\_test3, respectively. Source experiment data files offer subset-specific versions of train/validation/test in which the transcript may be the alternate forms of ASR or ASR-corrected versions. The naming convention prefix of these is according to the pattern: \{split\}\_\{subset\}\_\{transcript-version\}. Therefore, for example, train\_virtscribe\_humantrans.csv will give the training data from the \virtscribe{} subset with the original human transcription version; whereas  train\_virtscribe\_asr.csv will give the ASR transcription version.

\subsection{Metadata files (*\_metadata.csv)}
Metadata files include columns for the dataset name (e.g. \virtassist{}, \virtscribe{}, \aci{}), \textit{id}, \textit{encounter\_id}, \textit{doctor\_name}, \textit{patient\_firstname}, \textit{patient\_familyname}, \textit{gender}, \textit{chief complaint (cc)}, and \textit{secondary complaints (2nd\_complaints)}. Both \textit{id} and \textit{encounter\_id} can be used to identify a unique encounter. The \textit{encounter\_id} were the identifiers used for the MEDIQA-CHAT and MEDIQASUM 2023 competitions. The \textit{id} unique identifier will also denote a specific subset.

\subsection{Transcript/Note files (*.csv)}
In the source-target data files, transcript and note text are  given along with the dataset name and \textit{id} or \textit{encounter\_id}. This file may be joined with the metadata files using either \textit{id} or \textit{encounter\_id}. \textit{encounter\_id} should be used for challenge data, whereas the \textit{id} should be used for the source experiment data.

\section{Technical Validation}

\subsection{Content validation}
After dataset creation and cleaning, an additional content validation step was conducted to ensure medical soundness. For each encounter, medical annotators were tasked with reviewing each symptom, test, diagnosis and treatment from the encounter. In cases where the medical annotation specialist was unsure of certain facts (e.g. can drug X be prescribed at the same time as drug Y?), the encounter undergoes two possible additional reviews. Firstly, if the phenomenon in question can be searched identified from a +3M store of propriety clinical notes (which we will refer to as the \consult{} dataset)\footnote{These include notes from multiple provider organizations across the country; and as part of a remote scribe service.}, we deemed the information credible. Alternatively, if the information is not something that could be identified by the first approach, the question is escalated to a clinical expert annotator. Encounters with unexplainable or severe logical or medical problems identified by a medical annotators were removed (e.g. using a medication for urinary tract infection for upper respiratory infection).

\subsection{Comparison with real data}

To study differences between the \acidemo{} dataset and a set of real encounters, we conduct statistical comparison with 163 randomly chosen family medicine clinical encounters (including pairs human transcriptions and corresponding clinical notes) with in-depth alignment annotation, from the \consult{} dataset. Tables \ref{tab:datastatisitcs-real-vs-aci} and \ref{tab:alignment-stats} show the statistical comparison between the 20 encounters in the validation set (aci-validation) and the \consult{} encounters. In general, the \acidemo{} dataset had on average shorter notes, at 492 tokens versus 683 tokens for the consult dataset. Except for the \results{} division, every division was longer in the consult data (Table \ref{tab:datastatisitcs-real-vs-aci}). The \acidemo{} dataset also exhibits shorter dialogue lengths, by approximately 100 tokens and 20 sentences; as well a shorter notes by approximately 100 tokens (Table \ref{tab:alignment-stats}). One reason for the shorter note length is our removal of unsupported note text.

\begin{table}[ht]
\centering
\begin{tabular}{l|l|l|l|l|l}
\hline
& \subjective{} & \exam{} & \results{} & \assessment{} & full \\
\hline
 \textbf{consult} & & & & & \\
\hline
avg length (tok) & 393 & 149 & 19 & 122 & 683 \\
avg length (sentences) & 35 & 19 & 2 & 11 & 66\\
\hline
  \textbf{aci-validation} & & & & & \\
\hline
avg length (tok) &  229 & 48 & 23 & 192 & 492 \\
avg length (sentences) & 24 & 8 & 4 & 16 & 49 \\
\hline
\end{tabular}
\caption{\label{tab:datastatisitcs-real-vs-aci} Data statistic comparing notes from aci-validation with a sample of real doctor-patient. }
\end{table}

We additionally annotated for alignments of data between the source and target on the validation set (20 encounters) and consult set, similar to that of previous work\cite{yim-etal-2020-alignment}. This annotation marks associations between note sentences and their corresponding source transcript sentences. Unmarked note sentences indicate that a sentence may be purely structural (e.g. section header) or may include unsupported content. Likewise, unmarked transcript sentences may indicate that the content is superfluous. Comparing the portions of annotated alignments in separate corpora gives indications of corpora similarity with respect to relative content transfer. Other useful metrics which provide measures of alignment/generation difficulty include : (a) the fraction of alignment crossings (whether content appear monotonically versus "out-of-order"/"crossing") \cite{Tiedemann2011BitextA}), (b) the similarity of corresponding text segments, and (c) percentage of transcript speech modes. The results of these comparisons are shown in Table \ref{tab:alignment-stats}.

Labeled alignment annotations show that approximately the same fractions of dialogue and note sentences were labeled (0.34 and 0.49 transcript, 0.84 and 0.95 note for the consult and \acidemo{} corpus respectively); with a high 0.95 fraction for the \acidemo{} corpus, as designed by the removal of unsupported text. With shorter transcripts (1203 tokens in \acidemo{} vs 1505 tokens in the \consult{} set), the \acidemo{} corpus also had a 15\% more aligned transcript sentences. The text similarity (Jaccard unigram) of alignments were similar (0.15 and 0.12) as was the fraction of crossing annotations (0.67 and 0.95) for the \consult{} and \acidemo{} corpus respectively; though the dialogue-note document similarity was higher in the \acidemo{} corpus. 

\begin{table}[ht]
\centering
\begin{tabular}{l|l|l}
\hline
 & consult & aci-corpus \\
 \hline
\underline{dialogue} &  &  \\
avg length (no speaker tokens) (tok)  & 1505 & 1203 \\
avg length (sentences)  & 141 & 80 \\
\underline{note} &  &  \\
avg length (tok)  & 683 & 492 \\
avg length (sentences)  & 66 & 49 \\
\hline
\underline{annotation} &  &  \\
fraction note sentences aligned & 0.84 & 0.95 \\
fraction transcript sentences aligned & 0.34 & 0.49 \\
fraction crossing annotations & 0.67 & 0.75 \\
avg alignment text similarity & 0.15 & 0.12 \\
avg encounter dialogue-note text similarity & 0.26 & 0.31 \\
\hline
\underline{\% note sentences with labels} &  &  \\
DICTATION & 8 & 4 \\
QA & 15 & 43 \\
STATEMENT & 23 & 29 \\
STATEMENT2SCRIBE & 17 & 7 \\
\hline
\end{tabular}
\caption{\label{tab:alignment-stats}Alignment statistic comparison of aci-validation with a sample of real doctor-patient. }
\end{table}

The percentage of note sentences annotated with different labels\footnote{DICTATION: besides punctuation and formatting word-for-word copy-paste statements from the transcript, QA: question-answer conversation adjacency pairs, STATEMENT: conversation statements, STATEMENT2SCRIBE: directed instructions or content to a external scribe.} show across the board lower percentages in the \consult{} data. This is explainable as the transcript length and thus the percentage of note sentences annotated with a certain label will decrease. However, it is interesting to show that the \acidemo{} corpus had a higher percentage of note sentences coming from question-answer paired transcript sentences and conversation statements rather than dictation/statement2scribe. For example while in the \consult{} dataset, important QA makes up twice as much transcript sentences as in dictation (15\% and 8\%), in the \acidemo{} dataset there are ten times more QA labeled sentences than dictation (43\% vs 4\%). Meanwhile in the \consult{} dataset, transcript sentences identified with an alignment using the ``statement'' tag was about three times that of dictation, however this was about seven times in the \acidemo{} corpus. Together, this data suggests that the \acidemo{} corpus may be slightly less challenging in terms of documents lengths and has a skew towards question-answer and statements information content; though the magnitudes in lengths and similarity are comparable.

\subsection{Baseline experiments}
In this section, we present our baseline experiments designed to benchmark the \acidemo{} Corpus. These experiments encompass various note-generation tasks and incorporate state-of-the-art note-generation techniques. To assess the robustness of note-generation techniques, we also examine the impact of different clinical doctor-patient dialogue transcript generation methods with and without human correction on the quality of automatically generated clinical notes derived from these transcripts.

\subsubsection{Note generation models}
\label{models}
The experiments on note-generation models to benchmark the \acidemo{} Corpus are listed below:

\begin{description}

   \item[Transcript-copy-and-paste] Previous research finds taking the longest sentence \cite{longest_sentence_baseline_Gliwa_2019} as dialogue summarization is a good baseline. In the spirit of this approach, we adopt several variations to generate the clinical note: (1) the longest speaker's turn, (2) the longest doctor's turn, (3) the first two and the last ten speaker's turns, (4) the first two and the last ten doctors turns and (5) the entire transcript.

    \item[Retrieval-based] Borrowing from retrieval-based response generation \cite{jurafsky2008speech}, we pose a simple baseline that retrieves a relevant note in the training corpus rather than generating new text. To generate a clinical note for a new transcript, we employ transcript UMLS concept set similarity to retrieve the most similar transcript from the train set. The note that corresponds to this transcript in the training set is selected as the summarization for the new transcript, based on the assumption that the semantic overlap between the UMLS concepts in the two transcripts is a reliable indicator of their content similarity. Following the same manner, we adopt a similar retrieval-based method on the document embedding similarity from the spaCy English natural language process pipeline (https://spacy.io/).
   
   \item[BART-based] We employ the SOTA transformer model, bidirectional autoregressive transformer (BART) \cite{BART}. We also include its two variants: (1) a version with continued pre-training on PubMed abstract \cite{BioBART}, aimed at learning domain-specific language and knowledge, and (2) a version fine-tuned on the SAMSum corpus \cite{gliwa-etal-2019-samsum}, designed to enhance the model's performance on conversational summarization tasks. For all BART-based models, we use the BART-Large version. It is important to note that although BART and BioBART have the same model structure, they possess distinct tokenizers and vocabulary sizes. These differences play a significant role in determining their respective performance on the \acidemo{} corpus. The corresponding fine-tuning parameters can be found in the Appendix. BART-based models have the same limit of 1,024 tokens. 
   
   \item[LED-based] We leverage the Longformer-Encoder-Decoder (LED) architecture \cite{LED}, which incorporates an attention mechanism that can scale up to longer sentences. LED-based models have the same limit of 16K tokens. Because the transcript is long, LED overcomes the sentence length limit from BART. We also include its variant, which is finetuned on the Pubmed dataset \cite{PubMed_dataset}, to enhance the model's summarization ability in the biomedical context.  The corresponding fine-tuning parameters can be found in the Appendix.

   \item[OpenAI models] We experimented with the latest OpenAI models and APIs\footnote{\url{https://platform.openai.com/docs/models}}: (i) Text-davinci-002, (ii) Text-davinci-003, (iii) ChatGPT (gpt-3.5-turbo), and (iv) GPT-4. The first three models have the same limit of 4,097  tokens, shared between the prompt and the output/summary, whereas GPT-4 allows 32k tokens.  We used the following prompt: 
    \begin{itemize}
        \item Prompt: "summarize the conversation to generate a clinical note with four sections: HISTORY OF PRESENT ILLNESS, PHYSICAL EXAM, RESULTS, ASSESSMENT AND PLAN. The conversation is:"
    \end{itemize} 
   To allow adequate division detection, we added some light rule-based post-processing, adding endlines before and after for each section header. This post-processing described in Appendix Table \ref{tab:gpt-posprocess}.
   
\end{description}

\subsubsection{Full-note- vs division-based note-generation approaches}

In the cases of the fine-tuned pre-trained models, we investigated note generation with two overall approaches: full note generation versus division-based generation and concatenation. The first approach generates a complete note from the  transcript using a single model or approach. The latter approach is motivated by the long input and output lengths of our data -- which may exceed that of those pre-trained models are typically trained for. To this end, full notes were divided into the \subjective{}, \exam{}, \results{}, and \assessment{} divisions using a rule-based regular-expression section detection. As the notes were followed a handful of regular patterns, this section detection was highly performant. In cases where certain sections were missing, an \textit{EMPTY} flag was used as the output. Each division generation model was separately fine-tuned. The final note was created by concatenating the divisions.

\subsubsection{Automatic Evaluation Metrics}

We employ a variety of widely-used automatic evaluation metrics to evaluate performances in different perspectives. Specifically, we measure at least one lexical n-gram metric, an embedding-based similarity metric, a learned metric, and finally an information extraction metric. We evaluate the note generation performance both in the full note and in each division.

For the ngram-based lexical metric, we compute ROUGE\cite{lin2004rouge}(1/2/-L), which computes unigram, bigram, and the longest common subsequence matches between reference and candidate clinical notes. For an embedding-based metric, we applied BERTScore \cite{BERTSCORE} which greedily matches contextual token embeddings from pairwise cosine similarity. BERTScore efficiently captures synonym and context information. For a model-based learned metric, we used BLEURT\cite{BLUERT}, which is trained for scoring candidate-reference similarity. Additionally, we incorporate a medical concept- based evaluation metric (\medcon{}) to gauge the accuracy and consistency of clinical concepts. This metric calculates the F1-score to determine the similarity between the Unified Medical Language System (UMLS) concept sets in both candidate and reference clinical notes.\footnote{This is similar to the CheXpert evaluation for radiology summarization however our concepts are not restricted to 14 predetermined categories, and do not include weightings or assertion status.\cite{irvin2019chexpert}} The extraction of UMLS concepts within clinical notes is performed using a string match algorithm applied to the UMLS concept database through the QuickUMLS package\cite{soldaini2016quickumls}. To ensure clinical relevance, we restrict the \medcon{} metric to specific UMLS semantic groups, designated as \textit{Anatomy}, \textit{Chemicals} {\&}\textit{Drugs}, \textit{Device}, \textit{Disorders}, \textit{Genes} {\&} \textit{Molecular Sequences}, \textit{Phenomena} and  \textit{Physiology}. To consolidate the various evaluation metrics, we first take the average of the three ROUGE submetrics as ROUGE, and the average of ROUGE, BERTScore, BLEURT, and \medcon{} scores as the final evaluation score. Because BERTScore and BLEURT are limited by their pre-trained embedding length, we only use these evaluations for the division-based evaluation.

\subsubsection*{Results}

We fine-tune the models on the train set and select the best trained model based on evaluation on the validation set. Performances were evaluated on three test sets. Test sets 1 and 2 correspond to the test sets from  ACL ClinicalNLP MEDIQA-Chat 2023 TaskB full-note generation and TaskC dialogue generation, respectively. Test 3 corresponds to CLEF MEDIQA-SUM 2023 Subtask C full-note generation. 

Our test 1 full note evaluation results can be found in Tables \ref{tab:test1}. Per-division \subjective{}, \exam{}, \results{}, and \assessment{} results for test 1 are accounted for in Tables \ref{tab:test1_subjective}, \ref{tab:test1_objective_exam}, \ref{tab:test1_objective_results}, and \ref{tab:test1_assessment_and_plan}. In the main body of this paper, we discuss the results of test 1 which was used for our first full note generation task challenge\footnote{TaskB in https://github.com/abachaa/MEDIQA-Chat-2023}. We will first provide an overview of the model performance in both full-note and division-based evaluations. We will then describe each model type's performance. For reference, we provide the results of test 2 and test 3 in the Appendix.

In the full-note evaluation, the BART+FT$_{\mathrm{SAMSum}}$ (Division) model achieved the highest ROUGE scores, with 53.46 for ROUGE-1, 25.08 for ROUGE-2 and 48.62 for ROUGE-L. This is because when BART+FT$_{\mathrm{SAMSum}}$ (Division) model was fine-tuned on our \acidemo{} training set, it learned more specific clinical jargon in the \acidemo{} corpus, such as accurate subsection headers (``CHIEF COMPLAINT'', ``HISTORY OF PRESENT ILLNESS'', ...) and physical examination results (``- Monitoring of the heart: No murmurs, gallops ...'' ). On the contrary, GPT-4 demonstrated the highest \medcon{} evaluation score of 57.78, while achieving the second to third-best performance in ROUGE scores, with 51.76 for ROUGE-1, 22.58 for ROUGE-2 and 45.97 for ROUGE-L. The great performance can be attributed to the model's gigantic size, intensive pretraining, huge context size, and great versatility. GPT-4 captured many relevant clinical facts and thus had the highest \medcon{}. However, since it was not specifically fine-tuned for the \acidemo{} corpus clinical note format, it exhibited slightly inferior performance in capturing the structured \acidemo{} clinical notes. An example of a note generated from different models can be found in the Appendix Table \ref{tab:case_study}. Interestingly, the retrieval-based baselines showed very competitive ROUGE performances out-of-the-box with ROUGE-L of 40.47 F1 for and 38.20 F1 for the UMLS and sentence versions respectively. Furthermore, the simple transcript copy-and-paste baselines produced high starting points that out-performed untreated LED-based models. For example, simply copying the transcript achieved a 40.47 F1  ROUGE-L and 33.30 F1 \medcon{} score, whereas the fine-tuned division based LED model achieved 29.30 F1 and 32.67 F1.

\begin{table}[h]
\centering
\begin{tabular}{lcccc}
\hline
\textbf{Model}                           & \textbf{ROUGE-1} & \textbf{ROUGE-2} & \textbf{ROUGE-L} & \textbf{\medcon} \\ \hline
\textbf{Transcript-copy-and-paste} & & & &\\longest spearker turn&27.84&9.32&23.44&32.37\\ 
longest doctor turn&27.47&9.23&23.20&32.33\\ 
12 speaker turns&33.16&10.60&30.01&39.68\\ 
12 doctor turns&35.88&12.44&32.72&47.79\\ 
transcript&32.84&12.53&30.61&55.65\\ \cline{1-1}
\textbf{Retrieval-based} & & & &\\train$_{\mathrm{UMLS}}$&43.87&17.55&40.47&33.30\\ 
train$_{\mathrm{sent}}$&41.59&15.50&38.20&26.17\\ \cline{1-1}
\textbf{BART-based} & & & &\\BART&41.76&19.20&34.70&43.38\\ 
BART (Division)&51.56&24.06&45.92&47.23\\ 
BART+FT$_{\mathrm{SAMSum}}$&40.87&18.96&34.60&41.55\\ 
BART+FT$_{\mathrm{SAMSum}}$ (Division)&\textbf{53.46}&\textbf{25.08}&\textbf{48.62}&48.23\\ 
BioBART&39.09&17.24&33.19&42.82\\ 
BioBART (Division)&49.53&22.47&44.92&43.06\\ \cline{1-1}
\textbf{LED-based} & & & &\\LED&28.37&5.52&22.78&30.44\\ 
LED (Division)&34.15&8.01&29.80&32.67\\ 
LED+FT$_{\mathrm{PubMed}}$&27.19&5.30&21.80&27.44\\ 
LED+FT$_{\mathrm{PubMed}}$ (Division)&30.46&6.93&26.66&32.34\\ \cline{1-1}
\textbf{OpenAI (wo FT)} & & & &\\Text-Davinci-002&41.08&17.27&37.46&47.39\\ 
Text-Davinci-003&47.07&22.08&43.11&57.16\\ 
ChatGPT&47.44&19.01&42.47&55.84\\ 
GPT-4&51.76&22.58&45.97&\textbf{57.78}\\ 

\hline
\end{tabular}
\caption{Results of the summarization models evaluated at the full note level, test set 1. Simple retrieval-based methods provided strong baselines wih better out-of-the-box performances than LED models and full-note BART models.  In general for BART and LED fine-tuned models, division-based generation worked better. OpenAI models with simple prompts were shown to give competitive outputs despite no additional fine-tuning or dynamic prompting.}
\label{tab:test1}
\end{table}

In division-based evaluations, we found that different models achieved the highest average score across different note divisions,   BART+FT$_{\mathrm{SAMSum}}$ (Division) scored 51.08 in the \subjective{} division (Table \ref{tab:test1_subjective}), Text-davinci-003 reached 55.30, 48.90 and 46.19 in the \exam{} (Table \ref{tab:test1_objective_exam}), \results{} (Table \ref{tab:test1_objective_results}), and \assessment{} (Table \ref{tab:test1_assessment_and_plan}) divisions, respectively. These results indicate that all three models can be good candidates for the note-generation task. However, since BART+FT$_{\mathrm{SAMSum}}$ (Division) required fine-tuning and Text-davinci-003 did not, the latter two models demonstrated greater potential. A few additional examples for Text-davinci-003 could potentially enhance their performance, by enabling the models to learn specific clinical jargon in each division.

In comparing the full-note and division-based note-generation approaches, our experiments demonstrated that, for our pretrained BART- and LED-based models, division-based note-generation methods resulted in significant improvements over full-note-generation methods. These improvements ranged from 1 to 14 point increases in both ROUGE and \medcon{} evaluations for the full-note-based evaluation. This finding implies that breaking down a complex summarization problem into smaller divisions effectively captures more critical information. For division-based evaluations, the increase is not obvious for the \subjective{} divisions, but around 20 percent in the average score for \exam{}, \results{} and \assessment{} divisions. This can be attributed to the generation of the latter three divisions at the end of clinical notes, which often exceeds the word length of typical summarization tasks that BART-based and LED-based models are used for. Additionally, since some notes in the training set lack these divisions, the note-generation models struggle to learn the division structure during fine-tuning from the full note. As the division of clinical notes is identified by a rule-based division header extraction method, even when the information from a specific division is generated as a few sentences, the corresponding division information cannot be detected by the evaluation program.

\begin{table}[h]
\centering
\begin{tabular}{lccccccc}
\hline
\textbf{} & \multicolumn{7}{c}{\textbf{Evaluation score on the \subjective{} division}} \\ \cline{2-8} 
\textbf{Model} & \textbf{ROUGE-1} & \textbf{ROUGE-2} & \textbf{ROUGE-L} & \textbf{BERTScore} & \textbf{BLEURT} & \textbf{\medcon} & \textbf{Average} \\ \hline
\textbf{Retrieval-based} &  &  &  &  &  &  &\\ 
train$_{\mathrm{UMLS}}$&41.70&23.45&31.64&72.10&39.01&23.04&41.60\\ 
train$_{\mathrm{sent}}$&41.12&20.62&29.20&70.78&37.94&18.86&39.47\\ \cline{1-1}
\textbf{BART-based }&  &  &  &  &  &  &\\ 
BART&48.19&25.81&30.13&68.93&43.83&44.41&47.97\\ 
BART (Division)&47.25&26.05&31.21&70.05&43.55&44.20&48.16\\ 
BART+FT$_{\mathrm{SAMSum}}$&46.33&25.52&29.88&68.68&\textbf{45.01}&43.21&47.70\\ 
BART+FT$_{\mathrm{SAMSum}}$ (Division)&\textbf{52.44}&\textbf{30.44}&\textbf{35.83}&\textbf{72.41}&44.51&\textbf{47.84}&\textbf{51.08}\\ 
BioBART&45.79&23.65&28.96&68.49&41.09&41.10&45.87\\ 
BioBART (Division)&46.29&25.99&32.43&70.30&42.99&41.14&47.33\\ \cline{1-1}
\textbf{LED-based }&  &  &  &  &  &  &\\ 
LED&24.81&5.29&11.00&55.60&30.68&20.19&30.04\\ 
LED (Division)&31.27&8.31&15.99&56.94&25.40&24.03&31.22\\ 
LED+FT$_{\mathrm{PubMed}}$&23.48&4.72&10.49&54.46&20.32&17.91&26.40\\ 
LED+FT$_{\mathrm{PubMed}}$ (Division)&26.03&6.17&12.93&56.41&19.19&20.46&27.78\\ \cline{1-1}
\textbf{OpenAI (wo FT)} &  &  &  &  &  &  &\\ 
Text-Davinci-002&29.73&12.38&20.13&58.98&36.70&32.47&37.22\\ 
Text-Davinci-003&33.29&15.24&23.76&60.63&38.06&36.14&39.73\\ 
ChatGPT&32.70&14.05&22.69&65.14&39.48&38.21&41.49\\ 
GPT-4&41.20&19.02&26.56&63.34&43.18&44.25&44.93\\ 

\hline
\end{tabular}
\caption{Results of the summarization models on the \subjective{} division, test set 1. BART-based models generated at both full note and division levels had similar levels of performances, which were in general better than the other model classes. As in the full note evaluation, retrieval-based methods provided competitive baselines.}
\label{tab:test1_subjective}
\end{table}

\begin{table}[h]
\centering
\begin{tabular}{lccccccc}
\hline
\textbf{} & \multicolumn{7}{c}{\textbf{Evaluation score on the \exam{} division}} \\ \cline{2-8} 
\textbf{Model} & \textbf{ROUGE-1} & \textbf{ROUGE-2} & \textbf{ROUGE-L} & \textbf{BERTScore} & \textbf{BLEURT} & \textbf{\medcon} & \textbf{Average} \\ \hline
\textbf{Retrieval-based} &  &  &  &  &  &  &\\ 
train$_{\mathrm{UMLS}}$&43.43&25.77&36.74&\textbf{74.63}&40.96&24.63&43.88\\ 
train$_{\mathrm{sent}}$&37.02&19.58&31.20&71.47&35.83&14.52&37.77\\ \cline{1-1}
\textbf{BART-based }&  &  &  &  &  &  &\\ 
BART&0.56&0.36&0.56&40.25&10.66&0.00&12.85\\ 
BART (Division)&49.77&31.63&38.92&73.75&44.19&34.80&48.21\\ 
BART+FT$_{\mathrm{SAMSum}}$&6.22&3.74&5.21&44.33&14.82&4.14&17.09\\ 
BART+FT$_{\mathrm{SAMSum}}$ (Division)&47.73&29.51&36.98&73.41&42.86&35.91&47.56\\ 
BioBART&2.57&1.04&1.68&42.10&12.38&1.22&14.36\\ 
BioBART (Division)&42.51&26.15&32.19&71.57&42.18&29.55&44.23\\ \cline{1-1}
\textbf{LED-based }&  &  &  &  &  &  &\\ 
LED&0.00&0.00&0.00&0.00&14.87&0.00&3.72\\ 
LED (Division)&27.03&7.96&16.88&54.48&14.47&18.84&26.27\\ 
LED+FT$_{\mathrm{PubMed}}$&0.00&0.00&0.00&0.00&14.87&0.00&3.72\\ 
LED+FT$_{\mathrm{PubMed}}$ (Division)&20.24&6.30&12.14&54.13&12.67&18.07&24.44\\ \cline{1-1}
\textbf{OpenAI (wo FT)} &  &  &  &  &  &  &\\ 
Text-Davinci-002&43.68&22.31&35.03&68.25&45.68&35.41&45.75\\ 
Text-Davinci-003&\textbf{54.17}&\textbf{32.42}&\textbf{44.54}&73.40&\textbf{51.29}&\textbf{52.79}&\textbf{55.30}\\ 
ChatGPT&49.44&27.29&38.60&71.39&49.39&48.95&52.04\\ 
GPT-4&50.11&28.20&40.43&71.79&51.11&42.59&51.27\\ 

\hline
\end{tabular}
\caption{Results of the summarization models on the \exam{} division, test set 1. BART and LED full note generation models suffered a significant drop at the \exam{}. This may be attributable to the lower amounts of content required to be generated, the appearance of text later in the sequence, as well as the higher variety of structures. The OpenAI were in general better performant with BART division-based models as next best.}
\label{tab:test1_objective_exam}
\end{table}

\begin{table}[h]
\centering
\begin{tabular}{lccccccc}
\hline
\textbf{} & \multicolumn{7}{c}{\textbf{Evaluation score on the \results{} division}} \\ \cline{2-8} 
\textbf{Model} & \textbf{ROUGE-1} & \textbf{ROUGE-2} & \textbf{ROUGE-L} & \textbf{BERTScore} & \textbf{BLEURT} & \textbf{\medcon} & \textbf{Average} \\ \hline
\textbf{Retrieval-based} &  &  &  &  &  &  &\\ 
train$_{\mathrm{UMLS}}$&30.26&14.89&29.87&66.24&37.25&8.91&34.35\\ 
train$_{\mathrm{sent}}$&40.52&18.21&38.87&\textbf{73.33}&45.79&12.45&41.03\\ \cline{1-1}
\textbf{BART-based }&  &  &  &  &  &  &\\ 
BART&0.00&0.00&0.00&0.00&5.45&0.00&1.36\\ 
BART (Division)&30.48&19.16&27.80&66.64&43.07&21.56&39.27\\ 
BART+FT$_{\mathrm{SAMSum}}$&20.79&0.46&20.67&54.54&28.32&0.77&24.40\\ 
BART+FT$_{\mathrm{SAMSum}}$ (Division)&29.45&18.01&26.63&66.43&40.75&20.17&38.01\\ 
BioBART&17.50&0.00&17.50&52.44&25.33&0.00&22.36\\ 
BioBART (Division)&35.38&14.33&32.79&68.40&47.63&15.69&39.81\\ \cline{1-1}
\textbf{LED-based }&  &  &  &  &  &  &\\ 
LED&0.00&0.00&0.00&0.00&5.45&0.00&1.36\\ 
LED (Division)&14.04&4.97&11.08&48.86&9.61&7.86&19.09\\ 
LED+FT$_{\mathrm{PubMed}}$&0.00&0.00&0.00&0.00&5.45&0.00&1.36\\ 
LED+FT$_{\mathrm{PubMed}}$ (Division)&10.48&3.64&8.32&42.43&7.13&8.86&16.48\\ \cline{1-1}
\textbf{OpenAI (wo FT)} &  &  &  &  &  &  &\\ 
Text-Davinci-002&41.48&20.12&39.95&70.61&50.79&24.42&44.92\\ 
Text-Davinci-003&\textbf{44.92}&\textbf{25.21}&\textbf{43.84}&72.35&\textbf{55.87}&\textbf{29.37}&\textbf{48.90}\\ 
ChatGPT&34.50&17.75&30.84&66.68&48.51&22.28&41.29\\ 
GPT-4&37.65&19.94&35.73&68.33&48.50&26.73&43.67\\ 

\hline
\end{tabular}
\caption{Results of the summarization models on the \results{} division, test set 1. Similar to \exam{}, BART and LED full note generation models suffered a significant drop at the \results{} division. This may be attributable to the higher sparsity of this division, low amounts of content (sometimes only 2-3 sentences), and the appearance of text later in the sequence. The OpenAI were in general better performant with BART division-based models as next best.}
\label{tab:test1_objective_results}
\end{table}

\begin{table}[h]
\centering
\begin{tabular}{lccccccc}
\hline
\textbf{} & \multicolumn{7}{c}{\textbf{Evaluation score on the \assessment{} division}} \\ \cline{2-8} 
\textbf{Model} & \textbf{ROUGE-1} & \textbf{ROUGE-2} & \textbf{ROUGE-L} & \textbf{BERTScore} & \textbf{BLEURT} & \textbf{\medcon} & \textbf{Average} \\ \hline
\textbf{Retrieval-based} &  &  &  &  &  &  &\\ 
train$_{\mathrm{UMLS}}$&\textbf{44.59}&\textbf{21.50}&\textbf{29.66}&\textbf{70.39}&44.77&24.70&42.94\\ 
train$_{\mathrm{sent}}$&41.28&19.73&28.02&69.48&43.18&18.79&40.28\\ \cline{1-1}
\textbf{BART-based }&  &  &  &  &  &  &\\ 
BART&0.00&0.00&0.00&0.00&29.05&0.00&7.26\\ 
BART (Division)&43.31&20.59&26.55&67.49&40.99&32.30&42.73\\ 
BART+FT$_{\mathrm{SAMSum}}$&1.52&0.49&0.87&35.38&19.79&1.00&14.28\\ 
BART+FT$_{\mathrm{SAMSum}}$ (Division)&43.89&21.37&27.56&68.09&41.96&31.33&43.08\\ 
BioBART&0.00&0.00&0.00&0.00&29.05&0.00&7.26\\ 
BioBART (Division)&42.44&19.44&26.42&67.57&43.88&31.12&43.00\\ \cline{1-1}
\textbf{LED-based }&  &  &  &  &  &  &\\ 
LED&0.00&0.00&0.00&0.00&29.05&0.00&7.26\\ 
LED (Division)&28.23&6.13&12.44&55.75&27.78&21.94&30.27\\ 
LED+FT$_{\mathrm{PubMed}}$&0.00&0.00&0.00&0.00&29.05&0.00&7.26\\ 
LED+FT$_{\mathrm{PubMed}}$ (Division)&28.00&5.99&13.07&55.68&20.95&25.01&29.33\\ \cline{1-1}
\textbf{OpenAI (wo FT)} &  &  &  &  &  &  &\\ 
Text-Davinci-002&30.90&12.27&21.44&61.01&44.98&35.04&40.64\\ 
Text-Davinci-003&35.41&14.86&25.38&63.97&49.18&\textbf{46.40}&\textbf{46.19}\\ 
ChatGPT&36.43&12.50&23.32&63.56&48.21&43.71&44.89\\ 
GPT-4&38.16&14.12&24.90&64.26&\textbf{49.41}&42.36&45.44\\ 

\hline
\end{tabular}
\caption{Results of the summarization models on the \assessment{} division, test set 1. Similar to \exam{} and \results{}, BART and LED full note generation models suffered a significant drop at the \results{} division. This may be attributable to the appearance of text later in the sequence. The OpenAI were in general better performant with BART division-based models as next best.}
\label{tab:test1_assessment_and_plan}
\end{table}

\FloatBarrier

Our observations on the performance for each type of model are summarized below:

\begin{description}

   \item[Transcript-copy-and-paste] models are only evaluated in the full note. It demonstrated suboptimal performance, which is around 17 points less than the best ROUGE scores. This is primarily because transcripts from doctor-patient dialogues serve to facilitate doctor-patient interactions with questions, answers, and explanations related to various health phenomena. In contrast, clinical notes, which are created by and intended for healthcare professionals, generally follow the SOAP format to convey the information concisely and accurately. Therefore, transcripts and notes can differ significantly in terms of terminology, degree of formality, relevance to clinical issues, and the organization of clinical concepts. On the other hand, the original transcript often achieves the fourth highest score in \medcon{} evaluation at 55.65, owing to its ability to capture relevant UMLS concepts explicitly mentioned within the transcript.

    \item[Retrieval-based] models have the best BERTScore in \exam{}, \results{} and \assessment{} divisions in test 1, with around 1 to 5 points increase over the best BART-based and OpenAI models. They also have shown sometimes promising results with the second and first average scores in \results{} and \assessment{} divisions from test 3. This is because clinical notes with similar transcripts tend to have more similar clinical notes, especially when \results{} sections use standard phrasing and templates, or in scenarios where patients share common symptoms and health examinations from different medical problems. However, their performance in the \medcon{} evaluation metric is often poor because of the less accurate patient-specific medical conditions. As a result, these models may perform well in non-\medcon{} evaluation metrics but may not produce accurate \medcon{} evaluations.
   
   \item[BART-based] models demonstrated superior performance. In full-note evaluation,  BART+FT$_{\mathrm{SAMSum}}$ (Division) had the best ROUGE score performance with \medcon{} evaluation scores only secondary to OpenAI models. In \subjective{} division,  BART+FT$_{\mathrm{SAMSum}}$ (Division) had top performance in all scores except BLEURT. These findings suggest that using a model fine-tuned on a similar dataset serves as a solid foundation for summarization tasks. Meanwhile, BioBART exhibits a comparatively weaker performance than BART, which could be attributed to the choice of vocabularies, tokenizers, and consequently, the quality of contextual embeddings.  For BART-based models, the division-based note-generation approach improved the performance from the full-note-generation approach with around a 5 to 40 points increase in all division-based average scores. This implies that dividing the complex note-generation tasks into simpler subtasks can boost model performance.
   
   \item[LED-based] models were generally inferior to that of BART-based models with around 15 to 40 points lower scores in full-note ROUGE and \medcon{} scores. We observed that compared with the BART-based models, the LED-based models generate notes with worse fluency, less essential clinical information, and poorer division structure. On the other hand, the effect of the division method on LED-based models was similar to that on BART-based models, which lead to a 1 to 9 points increase in full-note ROUGE and \medcon{} scores and a 2 to 25 points increase in division-based average scores.

   \item[OpenAI] models exhibited good general performance and using a generic prompt, without fine-tuning. GPT4 outperformed other OpenAI models, at around 10 ROUGE-1 F1 points in full-note evaluation. This is consistent as GPT4 is known to have been trained with more parameters and has had shown to have made impressive performances across a variety of human tasks\cite{bubeck_sparks_2023,openai_gpt-4_2023}. While Text-davinci-003 and ChatGPT were within 4 ROUGE-1 points in test 1, there were larger 4-9 point gaps in test 2 and 3 respectively. This information combined with the relatively stable ROUGE-1 score for GPT4 (at around $\sim$50 Rouge-1), suggests that the earlier models had more unstable performances. Assessing the division-based performances, we see the relative ranking of the OpenAI's were more variable (with the exception of Text-davinci-002 consistently performing below the other models).
   
\end{description}

\subsubsection*{Effect of ASR vs human transcription and correction}

In practice, automatic speech recognition (ASR) is widely deployed because it provides an affordable, real-time text-based transcript. However, the quality of ASR is usually worse than the human transcript, influenced by its model type, hardware, and training corpus. To study the effect of ASR vs human transcription on clinical note generation from dialogue, we evaluate the note-general model performance on transcripts generated from these two approaches. We compare the performance between human transcription versus ASR for the \virtscribe{} subset of the data; and ASR versus ASR-corrected in the \aci{} subet. We conduct this ablation study with one of the best models in the previous section,  BART+FT$_{\mathrm{SAMSum}}$ (Division), and compare the result on the split of the three test sets. %We did not experiment with Open AI models, as fine-tuning-based models are more widely applied and are less robust in handling diverse input than Open AI models.

To study the difference with human transcription versus ASR for the \virtscribe{} subset, we experiment with feeding the raw ASR transcripts instead of it's original human transcription. We also fine-tune the model further to adapt to the ASR version by additionally learning for an additional 3 epochs with the same parameters using the ASR version of the \virtscribe{} train set. To understand effects of the train/decode discrepancies, we evaluate the results of feeding in the original human-transcription source as well as ASR versions to both the original fine-tuned model and the further ASR-fined tuned model.
%Because the train set is a combination of ASR with \virtscribe{} and correct ASR (ASRcor) with \aci{}, we further fine-tuned the BART+FT$_{\mathrm{SAMSum}}$ (Division) for 3 more epochs with the same parameter settings on each of ASR type as a subset of the train set respectively.
The results of the \virtscribe{} source experiments are presented in Table \ref{tab:ablation_asr_human}. We observed that the model setup with best ROUGE-1/2/L and \medcon{} scores are different for each test set. Namely, BART+FT$_{\mathrm{SAMSum}}$ (Division) with transcripts generated by ASR from \virtscribe{} dialogues do not exhibit outstanding differences with the human transcription (when using ASR source input performance dropped to 41.74 F1 ROUGE-L instead of 43.98 ROUGLE-L with the original model human transcript input for test1). Further fine-tuning  BART+FT$_{\mathrm{SAMSum}}$ (Division) with ASR notes in the train set also did not greatly improve the performance (fine-tuning improved 2 points in F1  to 43.82 F1 for an ASR transcript source, and a minimal drop to 43.59 when applying the original human transcription source). This indicates that ASR and the human transcript do not have a remarkable impact on the note-generation performance from  dialogue with \virtscribe{}.

\begin{table}[h]
\centering
\begin{tabular}{ccccccc}
\hline
\textbf{\begin{tabular}[c]{@{}c@{}}Test\\ set\end{tabular}} & \textbf{\begin{tabular}[c]{@{}c@{}}Bart\\ Fine-tuning\end{tabular}} & \textbf{\begin{tabular}[c]{@{}c@{}}Test\\ Split\end{tabular}} & \textbf{ROUGE-1} & \textbf{ROUGE-2} & \textbf{ROUGE-L} & \textbf{\medcon} \\ \hline
\multirow{4}{*}{1}&train&ASR&48.61&18.94&41.74&42.63\\ 
&+train$_{\mathrm{ASR}}$&ASR&\textbf{49.70}&19.96&43.82&41.96\\ 
&train&human&48.28&\textbf{20.09}&\textbf{43.98}&\textbf{46.13}\\ 
&+train$_{\mathrm{ASR}}$&human&48.50&19.52&43.59&42.85\\ 
\cline{1-3}\multirow{4}{*}{2}&train&ASR&\textbf{51.29}&\textbf{21.31}&43.76&\textbf{45.21}\\ 
&+train$_{\mathrm{ASR}}$&ASR&50.42&21.30&\textbf{44.68}&43.71\\ 
&train&human&50.11&20.80&44.44&43.35\\ 
&+train$_{\mathrm{ASR}}$&human&48.44&20.47&43.68&44.28\\ 
\cline{1-3}\multirow{4}{*}{3}&train&ASR&50.41&\textbf{20.01}&43.79&\textbf{49.91}\\ 
&+train$_{\mathrm{ASR}}$&ASR&49.22&19.72&43.19&44.18\\ 
&train&human&\textbf{50.86}&19.50&\textbf{44.59}&45.48\\ 
&+train$_{\mathrm{ASR}}$&human&47.42&18.42&42.67&44.72\\ 
\hline
\end{tabular}
\caption{Model performance on different test sets splits, comparison between \textit{virtscribe} dialogues with ASR and human transcript. The model finetuned on the train set is the BART+FT$_{\mathrm{SAMSum}}$ (Division) fine-tuned with 10 epochs on the original train set, as in the baseline methods. The train + train$_{\mathrm{ASR}}$ model refers to the BART+FT$_{\mathrm{SAMSum}}$ (Division) finetuned for 3 more epochs on the \textit{virtscribe} with ASR split of the train set.}
\label{tab:ablation_asr_human}
\end{table}

To study the effect of ASR versus ASR-corrected, we conduct similar experiments for the \aci{} subset by substituting the original ASR transcripts with corrected versions. The results of these experiments are shown in Table \ref{tab:ablation_asr_asrcorr}, we also observed that the model setup with best ROUGE-1/2/L and \medcon{} scores are different for each test set. The ASR-corrected did not exhibit more outstanding improvement from the original ASR on the BART+FT$_{\mathrm{SAMSum}}$ (Division)'s note generation performance (with approximately 1 F1 point difference amongst all the test sets and evaluation versions and metrics). Further fine-tuning  BART+FT$_{\mathrm{SAMSum}}$ (Division) with ASRcorr notes in the train set also did not substantially change performance. This indicates that those ASR errors corrected by humans do not have a remarkable impact on the note generation performance.

\begin{table}[h]
\centering
\begin{tabular}{ccccccc}
\hline
\textbf{\begin{tabular}[c]{@{}c@{}}Test\\ set\end{tabular}} & \textbf{\begin{tabular}[c]{@{}c@{}}Bart\\ Fine-tuning\end{tabular}} & \textbf{\begin{tabular}[c]{@{}c@{}}Test\\ Split\end{tabular}} & \textbf{ROUGE-1} & \textbf{ROUGE-2} & \textbf{ROUGE-L} & \textbf{\medcon} \\ \hline
\multirow{4}{*}{1}&train&ASR&54.03&24.19&48.09&45.97\\ 
&+train$_{\mathrm{ASRcorr}}$&ASR&\textbf{54.10}&24.47&47.92&47.17\\ 
&train&ASRcorr&54.04&24.35&\textbf{48.19}&46.00\\ 
&+train$_{\mathrm{ASRcorr}}$&ASRcorr&54.04&\textbf{24.50}&47.85&\textbf{47.36}\\ 
\cline{1-3}\multirow{4}{*}{2}&train&ASR&51.62&23.05&46.01&45.31\\ 
&+train$_{\mathrm{ASRcorr}}$&ASR&53.14&24.47&47.43&45.34\\ 
&train&ASRcorr&51.56&23.13&45.97&\textbf{45.98}\\ 
&+train$_{\mathrm{ASRcorr}}$&ASRcorr&\textbf{53.44}&\textbf{24.51}&\textbf{47.66}&45.47\\ 
\cline{1-3}\multirow{4}{*}{3}&train&ASR&53.07&\textbf{23.53}&\textbf{47.73}&45.27\\ 
&+train$_{\mathrm{ASRcorr}}$&ASR&52.53&23.14&47.06&\textbf{46.38}\\ 
&train&ASRcorr&\textbf{53.13}&23.48&47.63&45.52\\ 
&+train$_{\mathrm{ASRcorr}}$&ASRcorr&52.38&23.00&46.87&45.78\\ 
\hline
\end{tabular}
\caption{Model performance on different test sets splits, comparison between \textit{aci} dialogues with ASR and ASRcorr transcript. The model finetuned on the train set is the BART+FT$_{\mathrm{SAMSum}}$ (Division) fine-tuned with 10 epochs on the original train set, as in the baseline methods. The train + train$_{\mathrm{ASRcorr}}$ model refers to the BART+FT$_{\mathrm{SAMSum}}$ (Division) finetuned for 3 more epochs on the \textit{aci} with ASRcorr split of the train set.}
\label{tab:ablation_asr_asrcorr}
\end{table}

In summary, our investigation of ASR versus human transcription shows that although ASR can generate errors in the transcript, those errors do not have a remarkable impact on the note-generation performance and are thus tolerable by our current model setting. However, this could be due to our automatic evaluation metrics evaluating the n-grams and clinical facts with uniform weights. In clinical practice, some particular medical fact errors from the ASR can have a non-trivial impact.

\section{Usage Notes}

We have provided instructions in the README file in the Figshare repository describing how to process the \acidemo{} dataset. Examples of processing the data for different summarization evaluations can be found in the code located at the GitHub repository provided below.

\subsection{Limitations}
There are several limitations to this work. The data here is small and produced synthetically by medical annotators or patient actors in a single institution. Therefore, this dataset may not cover in a statistically representative way, all health topics, speech variations, and note format variations present in the real word. 

The data here is intended to be used for benchmarking methods related to clinician-patient dialogue summarization. It should not be used for training models to make medical diagnosis.

No patient data was used or disclosed here. Names of the original actors were changed. The gender balance of the entire dataset is roughly equal. Other demographic information was not modeled in this dataset.

\section{Code availability}
All code used to run data statistics, baseline models, and evaluation to analyze the \acidemo{} corpus is freely available at [LINK TO BE UPDATED]. 

\bibliography{references}  %%% Uncomment this line and comment out the ``thebibliography'' section below to use the external .bib file (using bibtex) .

\section{Author contributions statement}

WY developed and created the annotation guidelines, supervised the annotation work, advised on baseline experiments, performed corpus data analysis, and drafted the original manuscript. YF performed baseline experiments, analysis of model performance, and manuscript authorship. AB advised on guideline creation, annotation work, ran baselines, and reviewed and revised the manuscript. NS participated in acquisition of the source data, as well as advised on guideline creation, annotation work, and manuscript review. TL advised on baseline experiments, and reviewed and revised the manuscript. MY advised on the guideline creation, annotation work and baseline experiments, and reviewed and revised the manuscript. 

\section{Competing interests}

N/A

%%% Uncomment this section and comment out the \bibliography{references} line above to use inline references.
% \begin{thebibliography}{1}

% 	\bibitem{kour2014real}
% 	George Kour and Raid Saabne.
% 	\newblock Real-time segmentation of on-line handwritten arabic script.
% 	\newblock In {\em Frontiers in Handwriting Recognition (ICFHR), 2014 14th
% 			International Conference on}, pages 417--422. IEEE, 2014.

% 	\bibitem{kour2014fast}
% 	George Kour and Raid Saabne.
% 	\newblock Fast classification of handwritten on-line arabic characters.
% 	\newblock In {\em Soft Computing and Pattern Recognition (SoCPaR), 2014 6th
% 			International Conference of}, pages 312--318. IEEE, 2014.

% 	\bibitem{hadash2018estimate}
% 	Guy Hadash, Einat Kermany, Boaz Carmeli, Ofer Lavi, George Kour, and Alon
% 	Jacovi.
% 	\newblock Estimate and replace: A novel approach to integrating deep neural
% 	networks with existing applications.
% 	\newblock {\em arXiv preprint arXiv:1804.09028}, 2018.

% \end{thebibliography}

\section{Appendix}

\subsection{OpenAI post-processing}

\begin{table}[ht]
\centering
\begin{tabular}{p{.15\textwidth} p{.8\textwidth}}
\hline
 models & description\\
\hline
ChatGpt & \begin{lstlisting}[
    basicstyle=\footnotesize, %or \tiny or \footnotesize etc.
]
text = text.replace('\n', ' ').strip()
for division in [ 'HISTORY OF PRESENT ILLNESS', 'PHYSICAL EXAM', 
                                    'RESULTS', 'ASSESSMENT AND PLAN' ] :
        text = text.replace( '%s' %division, '\n%s\n' %division )
        text = text.replace( '<%s>' \%division, '\n%s\n' %division )
        text = text.replace( '%s:'\%division, '\n%s\n' %division )
        text = text.replace( '# %s:'\%division, '\n%s\n' %division )
        text = text.replace( '# %s'\%division, '\n%s\n' %division )
        text = text.replace( '## %s'\%division, '\n%s\n' %division )
        text = text.replace( '**%s**'\%division, '\n%s\n' %division )\end{lstlisting}\\
        \hline
Text-Davinci-002,Text-Davinci-003 & \begin{lstlisting}[basicstyle=\footnotesize]
text.replace('\n', ' ').strip().replace( 'PHYSICAL EXAM:', '\nPHYSICAL EXAM:\n' )
                .replace( 'RESULTS:', '\nRESULTS:\n' )
                .replace( 'ASSESSMENT AND PLAN:', '\nASSESSMENT AND PLAN:\n' )
\end{lstlisting}\\
        \hline
GPT-4 & \begin{lstlisting}[basicstyle=\footnotesize]
text = text.replace('\n', ' ').strip().
if (text.startswith("Possible summary:") or 
           text.startswith("Possible clinical note:") or 
           text.startswith("A possible clinical note is:")) : 
        text = text[text.index(":")+1:] 
text.replace( 'PHYSICAL EXAM:', '\nPHYSICAL EXAM:\n' )
                .replace( 'RESULTS:', '\nRESULTS:\n' )
                .replace( 'ASSESSMENT AND PLAN:', '\nASSESSMENT AND PLAN:\n' )
\end{lstlisting}\\
\hline
\end{tabular}
\caption{\label{tab:gpt-posprocess} Open-AI post-processing rules. In order to ensure the rule-based section algorithm may correctly split into divisions, we added several simple post-processing rules tailored to the algorithm.}
\end{table}

\subsection{Sample model output}

We investigated example outputs from different models, to generate notes from the transcript \textit{D2N080} in the validation set.  As demonstrated in Table \ref{tab:case_study}, both BART+FT$_{\mathrm{SAMSum}}$ (Division) and GPT-4 excelled at condensing dialogue information into a coherent clinical note. However, among all the models, only GPT-4 properly identified the patient's correction on the doctor's mistake in the transcript from ``right knee pain" to ``left knee pain". Meanwhile, BART+FT$_{\mathrm{SAMSum}}$ (Division) only missed crucial pain-related information and instead focused on less important details about the patient's travel.

\begin{table}
\centering
\resizebox{\columnwidth}{!}{%
\begin{tabular}{|cll|}
\hline
\multicolumn{2}{|c|}{\textbf{Transcript}} & \multicolumn{1}{c|}{\textbf{Note}} \\ \hline
\multicolumn{2}{|l|}{\begin{tabular}[c]{@{}l@{}}...\\ {[}doctor{]} i <understand> you're you've come in with some <right   knee pain> can you tell me about it what's going on\\ {[}patient{]} it it's not the <right knee> it's the <left   knee>\\ {[}doctor{]} okay the <\textbf{left knee}>\\ {[}patient{]} and it just happens occasionally less than once a day when i'm   <walking> all of a sudden it is kind of <like>\\ gives out and i <think> here i'm going to <fall> but i usually <catch>   myself so <lot> of times i have to hold a grocery\\ cart and that helps a   <lot> so it comes and goes and it it passes just about as quickly as it   comes i do n't know what it \\ is whether i stepped wrong or i just don't know...\end{tabular}} & \begin{tabular}[c]{@{}l@{}}<CHIEF COMPLAINT>    <\textbf{Left knee pain}>.\\ <HISTORY OF PRESENT ILLNESS>\\ Andrea Barnes is a 34-year-old <female> who \\ presents today for evaluation of <left knee pain>.\\ The patient has been experiencing intermittent \\ episodes of <pain> and sudden instability with \\ ambulation. Her <pain> is localized deep in her\\ <patella> and occurs less than once daily...\end{tabular} \\ \hline
\multicolumn{3}{|c|}{\textbf{Prediction}} \\ \hline
\multicolumn{1}{|c|}{\textbf{train$_{\mathrm{UMLS}}$}} & \multicolumn{1}{c|}{\textbf{BART+FT$_{\mathrm{SAMSum}}$}} & \multicolumn{1}{c|}{\textbf{BART+FT$_{\mathrm{SAMSum}}$   (Division)}} \\ \hline
\multicolumn{1}{|l|}{\begin{tabular}[c]{@{}l@{}}<CHIEF   COMPLAINT>            \textbf{Annual   exam}.\\ <HISTORY OF PRESENT ILLNESS>\\ Martha Collins is a 50-year-old <female> with a <past \\ medical history> significant for <congestive heart failure>, \\ <depression>,and <hypertension> who presents for her \\ annual exam. It has been a year since I last <saw> the patient...\end{tabular}} & \multicolumn{1}{l|}{\begin{tabular}[c]{@{}l@{}}<CHIEF COMPLAINT>         <\textbf{Right knee pain}>.\\ <HISTORY OF PRESENT ILLNESS>\\ The patient reports she has <right knee pain>, which she  \\ <experiences> occasionally less than once a day when \\ she is   <walking>. She does not know the cause of the \\ <pain>. She denies any specific <injury> to her <knee>...\end{tabular}} & \begin{tabular}[c]{@{}l@{}}CHIEF COMPLAINT>         <\textbf{Right knee pain}>.\\  <SOCIAL HISTORY>\\ <Patient> reports she <likes> to travel. She has a \\ <trip>  planned for the next month in which she \\ will <probably> be doing an <lot> of <walking>. \\ She is also <planning> to attend a   lion's club \\ <function> in which there are line dances. She does \\ not   <feel> her <knee> is in good <condition> for \\ this type of activity She denies <hearing> or \\ <feeling>  a pop at any point...\end{tabular} \\ \hline
\multicolumn{3}{|c|}{\textbf{Prediction}} \\ \hline
\multicolumn{1}{|c|}{\textbf{LED+FT$_{\mathrm{PubMed}}$}} & \multicolumn{1}{c|}{\textbf{LED+FT$_{\mathrm{PubMed}}$   (Division)}} & \multicolumn{1}{c|}{\textbf{GPT-4}} \\ \hline
\multicolumn{1}{|l|}{\begin{tabular}[c]{@{}l@{}}\textbf{CHKnee   <pain>} is a common <complaint> in the <emergency> \\ room today. The patient reports that it occurs less frequently than \\ previously thought and she does not know if her   <symptoms> \\ are caused by <walking> on <unstable> surfaces   or due to an \\ underlying <inflammatory disease> such  as <diabetes mellitus>, \\ <hypercholesterolemia> (hcm)...\end{tabular}} 
& 
\multicolumn{1}{l|}{\begin{tabular}[c]{@{}l@{}}CHHISTORY OF <PRESENT>   ILLNESS\\ Andrea <Diaz> is a pleasant 29-year old <female> who \\ presents to the clinic today for evaluation of <\textbf{right knee} \\ \textbf{pain}>. The onset and severity are <unknown> but they \\ usually resolve themselves quickly on their own without \\ treatment or therapy...\end{tabular}} 
&
\multicolumn{1}{l|}{\begin{tabular}[c]{@{}l@{}}<HISTORY OF PRESENT ILLNESS>: The patient \\ is a  65-year-old <female> who presents   with \\ intermittent <\textbf{left} \textbf{knee pain}> that occurs less than  \\ once a day  when <walking>. She describes it as a \\ <feeling>  of the <knee> giving out and the kneecap \\ fading. She   denies any <trauma>, pop, or <swelling>. \\ She has  not  taken any <analgesics>. She is concerned \\ about her  upcoming <trip>  that involves <walking> \\ and line <dancing>... \end{tabular}}

\\ \hline
\end{tabular}
}
\caption{Example outputs from different models, to generate notes from the transcript D2N080 in the validation set (reformatted). The UMLS concepts detected in fact-based evaluation are included inside angle brackets.}
\label{tab:case_study}
\end{table}

\subsection{Baseline hyper-parameters}

The fine-tuning and note generation hyper-parameters for BART- and LED-based baseline models can be found in Table \ref{tab:params}. Note that the max target token length is smaller than the total length of clinical notes. Because the BART- and LED-based baseline models are not initially pretrained with such a long token length as clinical notes, a longer max target token length does not have a very good generation result from our experiment.

\begin{table}
\centering
\begin{tabular}{ccc}
\hline
\textbf{Hyper-parameter} & \textbf{BART-Based} & \textbf{LED-Based} \\ \hline
Max source token length & 1024 & 2048 \\
Max target token length & 256 & 256 \\
Min target token length & 128 & 128 \\
Batch size & 1 & 2 \\
Epochs & 10 & 15 \\
Learning Rate & $10^{-5}$ & $10^{-5}$ \\
Weight decay & 0 & 0 \\
Beam size & 5 & 5 \\
Global attention & - & 128 \\ \hline
\end{tabular}
\caption{The fine-tuning and note generation hyper-parameters for BART- and LED-based baseline models.}
\label{tab:params}
\end{table}

\subsection{Results on test 2 and 3}

\begin{table}
\centering
\begin{tabular}{lcccc}
\hline
\textbf{Model}                           & \textbf{ROUGE-1} & \textbf{ROUGE-2} & \textbf{ROUGE-L} & \textbf{\medcon} \\ \hline
\textbf{Transcript-copy-and-paste} & & & &\\longest spearker turn&28.96&10.43&24.46&34.30\\ 
longest doctor turn&28.96&10.43&24.46&34.30\\ 
12 speaker turns&31.57&10.59&28.49&33.53\\ 
12 doctor turns&37.89&14.01&34.63&50.12\\ 
transcript&32.34&13.07&30.32&55.38\\ \cline{1-1}
\textbf{Retrieval-based} & & & &\\train$_{\mathrm{UMLS}}$&44.41&17.66&40.81&35.67\\ 
train$_{\mathrm{sent}}$&44.10&16.68&40.40&27.66\\ \cline{1-1}
\textbf{BART-based} & & & &\\BART&41.90&19.87&34.56&44.39\\ 
BART (Division)&\textbf{52.63}&\textbf{24.53}&46.71&46.97\\ 
BART+FT$_{\mathrm{SAMSum}}$&40.37&18.86&34.26&44.17\\ 
BART+FT$_{\mathrm{SAMSum}}$ (Division)&52.08&24.37&\textbf{47.16}&48.12\\ 
BioBART&39.00&18.44&33.40&43.05\\ 
BioBART (Division)&50.80&22.70&46.13&44.76\\ \cline{1-1}
\textbf{LED-based} & & & &\\LED&29.40&6.50&23.61&32.65\\ 
LED (Division)&35.14&8.57&30.84&34.24\\ 
LED+FT$_{\mathrm{PubMed}}$&27.66&6.13&22.31&31.98\\ 
LED+FT$_{\mathrm{PubMed}}$ (Division)&31.21&7.37&27.60&32.74\\ \cline{1-1}
\textbf{OpenAI (wo FT)} & & & &\\Text-Davinci-002&39.36&16.95&36.15&46.47\\ 
Text-Davinci-003&43.65&21.21&40.59&\textbf{55.92}\\ 
ChatGPT&42.30&16.57&37.31&49.50\\ 
GPT-4&51.24&21.65&45.60&55.04\\ 

\hline
\end{tabular}
\caption{Results of the summarization models at the full note level, test set 2. As in test 1, the most competitive models are the division-based BART models and GPT-4 in terms of Rouge scores. Unlike test 1, Text-Davinci-003 had a higher \medcon{} performance than GPT4.}
\label{tab:test2}
\end{table}

\begin{table}
\centering
\begin{tabular}{lccccccc}
\hline
\textbf{} & \multicolumn{7}{c}{\textbf{Evaluation score on the \subjective{} division}} \\ \cline{2-8} 
\textbf{Model} & \textbf{ROUGE-1} & \textbf{ROUGE-2} & \textbf{ROUGE-L} & \textbf{BERTScore} & \textbf{BLEURT} & \textbf{\medcon} & \textbf{Average} \\ \hline
\textbf{Retrieval-based} &  &  &  &  &  &  &\\ 
train$_{\mathrm{UMLS}}$&41.99&22.78&31.46&71.30&39.60&23.56&41.63\\ 
train$_{\mathrm{sent}}$&43.56&22.34&31.45&71.23&37.83&18.02&39.88\\ \cline{1-1}
\textbf{BART-based }&  &  &  &  &  &  &\\ 
BART&47.79&26.91&29.42&68.40&44.40&\textbf{47.93}&48.86\\ 
BART (Division)&50.41&28.98&32.72&70.14&45.44&47.40&50.09\\ 
BART+FT$_{\mathrm{SAMSum}}$&46.74&25.99&30.19&68.70&\textbf{45.76}&45.83&48.65\\ 
BART+FT$_{\mathrm{SAMSum}}$ (Division)&\textbf{50.43}&\textbf{30.11}&35.10&\textbf{71.79}&45.05&46.61&\textbf{50.50}\\ 
BioBART&48.37&26.72&31.47&69.25&43.18&45.36&48.33\\ 
BioBART (Division)&48.99&28.90&\textbf{35.14}&71.21&43.76&44.61&49.31\\ \cline{1-1}
\textbf{LED-based }&  &  &  &  &  &  &\\ 
LED&25.00&6.05&11.12&55.56&29.96&22.25&30.46\\ 
LED (Division)&31.21&8.55&16.12&57.06&28.08&24.21&31.99\\ 
LED+FT$_{\mathrm{PubMed}}$&23.20&5.37&10.50&54.97&22.04&19.20&27.31\\ 
LED+FT$_{\mathrm{PubMed}}$ (Division)&25.37&6.69&12.79&56.18&19.67&20.99&27.95\\ \cline{1-1}
\textbf{OpenAI (wo FT)} &  &  &  &  &  &  &\\ 
Text-Davinci-002&27.11&13.01&19.81&56.48&36.22&31.74&36.10\\ 
Text-Davinci-003&29.05&15.00&22.88&58.97&37.40&38.32&39.25\\ 
ChatGPT&27.65&12.40&18.92&59.55&38.37&32.17&37.44\\ 
GPT-4&40.40&18.72&26.81&63.13&43.59&46.98&45.58\\ 

\hline
\end{tabular}
\caption{Results of the summarization models on the \subjective{} division, test set 2. Similar to test 1, the best overall model proved to be BART+FT$_{\mathrm{SAMSum}}$.}
\label{tab:test2_subjective}
\end{table}

\begin{table}
\centering
\begin{tabular}{lccccccc}
\hline
\textbf{} & \multicolumn{7}{c}{\textbf{Evaluation score on the \exam{} division}} \\ \cline{2-8} 
\textbf{Model} & \textbf{ROUGE-1} & \textbf{ROUGE-2} & \textbf{ROUGE-L} & \textbf{BERTScore} & \textbf{BLEURT} & \textbf{\medcon} & \textbf{Average} \\ \hline
\textbf{Retrieval-based} &  &  &  &  &  &  &\\ 
train$_{\mathrm{UMLS}}$&44.48&26.70&36.60&\textbf{73.08}&40.80&23.52&43.33\\ 
train$_{\mathrm{sent}}$&41.14&21.58&33.60&72.43&38.35&18.07&40.24\\ \cline{1-1}
\textbf{BART-based }&  &  &  &  &  &  &\\ 
BART&2.99&0.07&2.99&40.87&14.28&0.00&14.29\\ 
BART (Division)&\textbf{48.36}&\textbf{28.19}&35.29&71.94&41.75&32.55&45.88\\ 
BART+FT$_{\mathrm{SAMSum}}$&2.83&0.26&2.83&40.53&14.36&0.00&14.22\\ 
BART+FT$_{\mathrm{SAMSum}}$ (Division)&46.52&28.00&34.87&71.81&40.63&31.94&45.21\\ 
BioBART&0.00&0.00&0.00&0.00&17.28&0.00&4.32\\ 
BioBART (Division)&42.31&23.90&30.42&70.45&39.43&28.18&42.57\\ \cline{1-1}
\textbf{LED-based }&  &  &  &  &  &  &\\ 
LED&0.00&0.00&0.00&0.00&17.28&0.00&4.32\\ 
LED (Division)&27.79&7.87&16.16&54.41&15.00&20.75&26.86\\ 
LED+FT$_{\mathrm{PubMed}}$&0.00&0.00&0.00&0.00&17.28&0.00&4.32\\ 
LED+FT$_{\mathrm{PubMed}}$ (Division)&21.05&5.85&11.53&54.01&13.99&16.46&24.32\\ \cline{1-1}
\textbf{OpenAI (wo FT)} &  &  &  &  &  &  &\\ 
Text-Davinci-002&38.73&19.19&30.53&65.51&43.69&39.93&44.65\\ 
Text-Davinci-003&47.30&27.30&\textbf{37.70}&69.44&47.69&\textbf{47.76}&\textbf{50.58}\\ 
ChatGPT&30.67&12.69&24.88&59.88&36.26&28.79&36.92\\ 
GPT-4&45.55&23.17&36.61&69.11&\textbf{49.13}&46.31&49.91\\ 

\hline
\end{tabular}
\caption{Results of the summarization models on the \exam{} division, test set 2. Similar to test 1, LED models found this division challenging to summarize with full note and the most performant model based on the averaged score is Text-davinci-003 with BART (Division) second best.}
\label{tab:test2_objective_exam}
\end{table}

\begin{table}
\centering
\begin{tabular}{lccccccc}
\hline
\textbf{} & \multicolumn{7}{c}{\textbf{Evaluation score on the \results{} division}} \\ \cline{2-8} 
\textbf{Model} & \textbf{ROUGE-1} & \textbf{ROUGE-2} & \textbf{ROUGE-L} & \textbf{BERTScore} & \textbf{BLEURT} & \textbf{\medcon} & \textbf{Average} \\ \hline
\textbf{Retrieval-based} &  &  &  &  &  &  &\\ 
train$_{\mathrm{UMLS}}$&\textbf{40.09}&14.85&\textbf{39.62}&71.37&46.40&13.13&40.61\\ 
train$_{\mathrm{sent}}$&39.18&16.08&38.44&\textbf{72.69}&43.36&7.25&38.63\\ \cline{1-1}
\textbf{BART-based }&  &  &  &  &  &  &\\ 
BART&22.50&0.00&22.50&56.07&30.48&0.00&25.39\\ 
BART (Division)&25.52&15.65&23.47&63.19&39.47&21.35&36.39\\ 
BART+FT$_{\mathrm{SAMSum}}$&0.00&0.00&0.00&0.00&6.95&0.00&1.74\\ 
BART+FT$_{\mathrm{SAMSum}}$ (Division)&22.97&13.74&21.27&62.14&37.45&15.34&33.56\\ 
BioBART&0.00&0.00&0.00&0.00&6.95&0.00&1.74\\ 
BioBART (Division)&23.38&12.66&21.88&61.99&39.44&16.35&34.27\\ \cline{1-1}
\textbf{LED-based }&  &  &  &  &  &  &\\ 
LED&0.00&0.00&0.00&0.00&6.95&0.00&1.74\\ 
LED (Division)&12.31&4.26&9.58&47.64&9.08&7.31&18.19\\ 
LED+FT$_{\mathrm{PubMed}}$&0.00&0.00&0.00&0.00&6.95&0.00&1.74\\ 
LED+FT$_{\mathrm{PubMed}}$ (Division)&8.85&2.88&7.26&42.02&6.30&7.63&15.57\\ \cline{1-1}
\textbf{OpenAI (wo FT)} &  &  &  &  &  &  &\\ 
Text-Davinci-002&34.17&15.76&32.82&67.08&47.62&19.25&40.38\\ 
Text-Davinci-003&36.73&\textbf{20.71}&36.12&67.73&\textbf{49.28}&22.64&\textbf{42.71}\\ 
ChatGPT&25.92&6.86&25.20&60.72&39.16&9.45&32.16\\ 
GPT-4&32.18&16.86&29.71&64.82&47.13&\textbf{25.24}&40.86\\ 

\hline
\end{tabular}
\caption{Results of the summarization models on the \results{} division, test set 2. Similar \exam{} results, LED models performed suboptimally for the objective\_results division. This could be as a result of minimal information content for this division (often empty) as well as this content appearing later in a long sequence. OpenAI models performed the best in this division.}
\label{tab:test2_objective_results}
\end{table}

\begin{table}
\centering
\begin{tabular}{lccccccc}
\hline
\textbf{} & \multicolumn{7}{c}{\textbf{Evaluation score on the \assessment{} division}} \\ \cline{2-8} 
\textbf{Model} & \textbf{ROUGE-1} & \textbf{ROUGE-2} & \textbf{ROUGE-L} & \textbf{BERTScore} & \textbf{BLEURT} & \textbf{\medcon} & \textbf{Average} \\ \hline
\textbf{Retrieval-based} &  &  &  &  &  &  &\\ 
train$_{\mathrm{UMLS}}$&\textbf{45.13}&\textbf{21.24}&\textbf{29.65}&\textbf{70.96}&43.72&25.30&43.00\\ 
train$_{\mathrm{sent}}$&42.10&20.02&28.27&70.09&42.56&16.93&39.93\\ \cline{1-1}
\textbf{BART-based }&  &  &  &  &  &  &\\ 
BART&0.79&0.26&0.57&34.86&19.88&0.00&13.82\\ 
BART (Division)&42.70&19.47&25.00&67.25&40.51&30.79&41.90\\ 
BART+FT$_{\mathrm{SAMSum}}$&1.19&0.45&0.65&35.07&20.20&0.54&14.14\\ 
BART+FT$_{\mathrm{SAMSum}}$ (Division)&42.59&19.58&25.61&67.66&41.12&32.22&42.56\\ 
BioBART&0.48&0.14&0.34&34.64&19.27&0.62&13.71\\ 
BioBART (Division)&41.96&19.05&25.59&66.95&41.15&28.58&41.39\\ \cline{1-1}
\textbf{LED-based }&  &  &  &  &  &  &\\ 
LED&0.00&0.00&0.00&0.00&29.99&0.00&7.50\\ 
LED (Division)&28.96&6.08&12.53&56.30&28.09&22.51&30.69\\ 
LED+FT$_{\mathrm{PubMed}}$&0.00&0.00&0.00&0.00&29.99&0.00&7.50\\ 
LED+FT$_{\mathrm{PubMed}}$ (Division)&29.47&6.33&12.80&55.98&21.21&24.29&29.42\\ \cline{1-1}
\textbf{OpenAI (wo FT)} &  &  &  &  &  &  &\\ 
Text-Davinci-002&30.66&11.84&19.38&59.95&44.54&34.33&39.86\\ 
Text-Davinci-003&34.91&15.88&25.06&63.25&47.74&\textbf{41.50}&44.44\\ 
ChatGPT&25.22&9.20&15.75&54.75&40.81&27.93&35.05\\ 
GPT-4&39.38&15.22&24.76&64.56&\textbf{49.78}&38.67&\textbf{44.86}\\ 

\hline
\end{tabular}
\caption{Results of the summarization models on the \assessment{}, test set 2. BART and LED models trained for full note generation perform suboptimally, likely as this content appearing later in a long sequence. Similar to test 1, the best models for \assessment{} are the BART Division models and the OpenAI Text-davinci-003 and GPT4 models. }
\label{tab:test2_assessment_and_plan}
\end{table}

\begin{table}
\centering
\begin{tabular}{lcccc}
\hline
\textbf{Model}                           & \textbf{ROUGE-1} & \textbf{ROUGE-2} & \textbf{ROUGE-L} & \textbf{\medcon} \\ \hline
\textbf{Transcript-copy-and-paste} & & & &\\longest spearker turn&25.37&9.05&21.85&29.30\\ 
longest doctor turn&25.13&9.04&21.79&29.60\\ 
12 speaker turns&31.00&10.72&28.68&36.60\\ 
12 doctor turns&34.69&12.51&32.27&45.42\\ 
transcript&32.75&12.78&30.91&\textbf{56.31}\\ \cline{1-1}
\textbf{Retrieval-based} & & & &\\train$_{\mathrm{UMLS}}$&48.00&20.57&44.61&38.99\\ 
train$_{\mathrm{sent}}$&40.86&14.66&37.64&25.24\\ \cline{1-1}
\textbf{BART-based} & & & &\\BART&40.54&18.52&34.62&44.92\\ 
BART (Division)&51.79&23.34&46.62&46.06\\ 
BART+FT$_{\mathrm{SAMSum}}$&39.38&18.38&33.89&46.01\\ 
BART+FT$_{\mathrm{SAMSum}}$ (Division)&\textbf{52.77}&\textbf{24.38}&\textbf{48.03}&47.56\\ 
BioBART&38.32&17.39&33.39&43.06\\ 
BioBART (Division)&50.28&22.95&46.09&43.21\\ \cline{1-1}
\textbf{LED-based} & & & &\\LED&28.96&5.80&23.66&33.47\\ 
LED (Division)&34.71&8.03&30.77&33.79\\ 
LED+FT$_{\mathrm{PubMed}}$&26.32&5.24&21.92&27.53\\ 
LED+FT$_{\mathrm{PubMed}}$ (Division)&31.07&7.52&27.83&33.74\\ \cline{1-1}
\textbf{OpenAI (wo FT)} & & & &\\Text-Davinci-002&41.02&18.93&38.50&49.05\\ 
Text-Davinci-003&42.57&21.13&39.89&54.93\\ 
ChatGPT&46.08&19.36&41.72&52.47\\ 
GPT-4&50.30&21.67&45.67&54.98\\ 

\hline
\end{tabular}
\caption{Results of the summarization models at the full note level, test set 3. As in test 1, the most competitive models are the division-based BART models and GPT-4 in terms of Rouge scores. Unlike test 1, the trancript-copy baseline had a highest \medcon{} performance than GPT4. However still the OpenAI models were the next best models in terms of both Rouge and \medcon{} scores.}
\label{tab:test3}
\end{table}

\begin{table}
\centering
\begin{tabular}{lccccccc}
\hline
\textbf{} & \multicolumn{7}{c}{\textbf{Evaluation score on the \subjective{} division}} \\ \cline{2-8} 
\textbf{Model} & \textbf{ROUGE-1} & \textbf{ROUGE-2} & \textbf{ROUGE-L} & \textbf{BERTScore} & \textbf{BLEURT} & \textbf{\medcon} & \textbf{Average} \\ \hline
\textbf{Retrieval-based} &  &  &  &  &  &  &\\ 
train$_{\mathrm{UMLS}}$&47.49&26.86&35.45&\textbf{73.83}&40.89&26.80&44.53\\ 
train$_{\mathrm{sent}}$&41.20&21.42&30.18&71.08&39.07&15.27&39.09\\ \cline{1-1}
\textbf{BART-based }&  &  &  &  &  &  &\\ 
BART&47.02&24.99&28.86&68.01&43.75&43.59&47.24\\ 
BART (Division)&48.92&27.14&31.77&70.15&\textbf{44.75}&45.12&48.99\\ 
BART+FT$_{\mathrm{SAMSum}}$&46.96&25.11&30.06&69.36&44.18&44.15&47.93\\ 
BART+FT$_{\mathrm{SAMSum}}$ (Division)&\textbf{52.35}&\textbf{29.96}&\textbf{35.60}&72.23&43.02&\textbf{45.77}&\textbf{50.08}\\ 
BioBART&46.77&24.70&30.13&68.68&42.06&39.58&46.05\\ 
BioBART (Division)&47.51&26.50&32.83&70.52&42.99&38.37&46.87\\ \cline{1-1}
\textbf{LED-based }&  &  &  &  &  &  &\\ 
LED&24.42&5.57&11.19&55.48&29.34&20.76&29.83\\ 
LED (Division)&31.28&8.59&16.08&57.59&27.66&24.15&32.01\\ 
LED+FT$_{\mathrm{PubMed}}$&22.47&4.99&10.07&54.53&20.13&16.35&25.88\\ 
LED+FT$_{\mathrm{PubMed}}$ (Division)&25.10&6.56&12.68&56.36&19.57&19.43&27.54\\ \cline{1-1}
\textbf{OpenAI (wo FT)} &  &  &  &  &  &  &\\ 
Text-Davinci-002&29.09&13.13&20.66&57.86&37.65&34.41&37.72\\ 
Text-Davinci-003&29.97&14.75&22.17&58.07&39.37&37.10&39.21\\ 
ChatGPT&30.62&13.70&21.33&62.26&38.30&35.38&39.46\\ 
GPT-4&39.33&17.75&25.54&61.97&41.59&42.62&43.43\\ 

\hline
\end{tabular}
\caption{Results of the summarization models on the \subjective{} division, test set 3. Similar to test 1, the best overall model proved to be BART+FT$_{\mathrm{SAMSum}}$. In this random sample, the train$_{\mathrm{UMLS}}$ model was competitive with some of the OpenAI baselines.}
\label{tab:test3_subjective}
\end{table}

\begin{table}
\centering
\begin{tabular}{lccccccc}
\hline
\textbf{} & \multicolumn{7}{c}{\textbf{Evaluation score on the \exam{} division}} \\ \cline{2-8} 
\textbf{Model} & \textbf{ROUGE-1} & \textbf{ROUGE-2} & \textbf{ROUGE-L} & \textbf{BERTScore} & \textbf{BLEURT} & \textbf{\medcon} & \textbf{Average} \\ \hline
\textbf{Retrieval-based} &  &  &  &  &  &  &\\ 
train$_{\mathrm{UMLS}}$&47.92&\textbf{31.98}&\textbf{41.42}&\textbf{75.01}&43.77&28.67&46.97\\ 
train$_{\mathrm{sent}}$&38.01&22.79&32.66&72.09&39.05&20.46&40.69\\ \cline{1-1}
\textbf{BART-based }&  &  &  &  &  &  &\\ 
BART&0.00&0.00&0.00&0.00&14.95&0.00&3.74\\ 
BART (Division)&45.55&27.45&33.89&71.52&42.17&30.91&45.06\\ 
BART+FT$_{\mathrm{SAMSum}}$&4.42&1.25&4.10&42.16&15.05&0.83&15.33\\ 
BART+FT$_{\mathrm{SAMSum}}$ (Division)&46.58&26.46&36.36&72.87&42.23&29.28&45.21\\ 
BioBART&5.63&1.96&5.27&43.58&15.88&2.29&16.51\\ 
BioBART (Division)&40.95&25.42&31.80&69.95&39.84&27.50&42.50\\ \cline{1-1}
\textbf{LED-based }&  &  &  &  &  &  &\\ 
LED&0.00&0.00&0.00&0.00&14.95&0.00&3.74\\ 
LED (Division)&27.38&8.98&16.23&54.19&16.18&19.76&26.91\\ 
LED+FT$_{\mathrm{PubMed}}$&0.00&0.00&0.00&0.00&14.95&0.00&3.74\\ 
LED+FT$_{\mathrm{PubMed}}$ (Division)&20.16&6.13&12.10&53.62&12.84&16.44&23.93\\ \cline{1-1}
\textbf{OpenAI (wo FT)} &  &  &  &  &  &  &\\ 
Text-Davinci-002&41.63&23.33&33.26&67.47&45.79&36.25&45.56\\ 
Text-Davinci-003&\textbf{49.39}&29.39&41.01&70.49&\textbf{48.96}&\textbf{46.24}&\textbf{51.40}\\ 
ChatGPT&44.06&23.72&34.92&68.05&47.22&40.89&47.60\\ 
GPT-4&44.20&24.07&36.64&68.58&47.87&39.36&47.70\\ 

\hline
\end{tabular}
\caption{Results of the summarization models on the \exam{} division, test set 3. Similar to test 1, LED models found this division challenging to summarize with full note and the most performant model based on the averaged score is Text-davinci-003. In this sample, BART+FT$_{\mathrm{SAMSum}}$ achieved second best performances.}
\label{tab:test3_objective_exam}
\end{table}

\begin{table}
\centering
\begin{tabular}{lccccccc}
\hline
\textbf{} & \multicolumn{7}{c}{\textbf{Evaluation score on the \results{} division}} \\ \cline{2-8} 
\textbf{Model} & \textbf{ROUGE-1} & \textbf{ROUGE-2} & \textbf{ROUGE-L} & \textbf{BERTScore} & \textbf{BLEURT} & \textbf{\medcon} & \textbf{Average} \\ \hline
\textbf{Retrieval-based} &  &  &  &  &  &  &\\ 
train$_{\mathrm{UMLS}}$&\textbf{37.30}&18.03&\textbf{35.82}&69.97&42.77&14.14&39.31\\ 
train$_{\mathrm{sent}}$&35.33&12.33&34.28&\textbf{70.50}&41.06&5.73&36.15\\ \cline{1-1}
\textbf{BART-based }&  &  &  &  &  &  &\\ 
BART&0.00&0.00&0.00&0.00&7.57&0.00&1.89\\ 
BART (Division)&25.97&16.01&23.19&62.86&39.77&19.60&35.99\\ 
BART+FT$_{\mathrm{SAMSum}}$&33.43&0.26&33.02&61.44&39.09&0.59&30.84\\ 
BART+FT$_{\mathrm{SAMSum}}$ (Division)&28.98&14.99&26.65&64.80&41.66&17.09&36.77\\ 
BioBART&32.80&0.21&32.80&60.93&38.95&0.00&30.45\\ 
BioBART (Division)&31.19&12.88&28.91&65.76&45.94&14.83&37.71\\ \cline{1-1}
\textbf{LED-based }&  &  &  &  &  &  &\\ 
LED&0.00&0.00&0.00&0.00&7.57&0.00&1.89\\ 
LED (Division)&12.73&4.19&9.20&48.87&10.80&6.61&18.75\\ 
LED+FT$_{\mathrm{PubMed}}$&0.00&0.00&0.00&0.00&7.57&0.00&1.89\\ 
LED+FT$_{\mathrm{PubMed}}$ (Division)&9.96&3.65&7.82&41.39&5.97&6.73&15.31\\ \cline{1-1}
\textbf{OpenAI (wo FT)} &  &  &  &  &  &  &\\ 
Text-Davinci-002&30.19&16.39&28.18&63.93&44.02&15.39&37.07\\ 
Text-Davinci-003&36.83&\textbf{21.14}&34.66&67.38&\textbf{50.28}&19.92&\textbf{42.11}\\ 
ChatGPT&30.66&12.92&27.89&63.72&45.31&15.36&37.05\\ 
GPT-4&32.63&18.57&30.21&63.88&45.68&\textbf{22.53}&39.81\\ 

\hline
\end{tabular}
\caption{Results of the summarization models on the \results{} division, test set 3. Similar \exam{} results, LED models performed suboptimally for the objective\_results division. This could be as a result of minimal information content for this division (often empty) as well as this content appearing later in a long sequence. OpenAI models performed the best in this division.}
\label{tab:test3_objective_results}
\end{table}

\begin{table}
\centering
\begin{tabular}{lccccccc}
\hline
\textbf{} & \multicolumn{7}{c}{\textbf{Evaluation score on the \assessment{} division}} \\ \cline{2-8} 
\textbf{Model} & \textbf{ROUGE-1} & \textbf{ROUGE-2} & \textbf{ROUGE-L} & \textbf{BERTScore} & \textbf{BLEURT} & \textbf{\medcon} & \textbf{Average} \\ \hline
\textbf{Retrieval-based} &  &  &  &  &  &  &\\ 
train$_{\mathrm{UMLS}}$&\textbf{47.13}&\textbf{23.72}&\textbf{31.41}&\textbf{71.52}&44.98&28.93&\textbf{44.88}\\ 
train$_{\mathrm{sent}}$&39.67&17.10&25.90&67.59&42.48&16.22&38.46\\ \cline{1-1}
\textbf{BART-based }&  &  &  &  &  &  &\\ 
BART&0.00&0.00&0.00&0.00&32.03&0.00&8.01\\ 
BART (Division)&43.99&19.44&25.89&66.83&42.30&31.98&42.72\\ 
BART+FT$_{\mathrm{SAMSum}}$&1.47&0.61&1.12&35.59&22.28&0.26&14.80\\ 
BART+FT$_{\mathrm{SAMSum}}$ (Division)&43.29&19.65&26.20&67.11&42.40&34.40&43.41\\ 
BioBART&1.10&0.78&1.01&35.32&21.72&0.85&14.71\\ 
BioBART (Division)&44.23&20.89&27.55&67.89&43.64&31.92&43.59\\ \cline{1-1}
\textbf{LED-based }&  &  &  &  &  &  &\\ 
LED&0.00&0.00&0.00&0.00&32.03&0.00&8.01\\ 
LED (Division)&28.53&5.57&12.36&55.66&27.74&20.72&29.90\\ 
LED+FT$_{\mathrm{PubMed}}$&0.00&0.00&0.00&0.00&32.03&0.00&8.01\\ 
LED+FT$_{\mathrm{PubMed}}$ (Division)&29.37&7.02&13.34&56.15&22.73&25.56&30.25\\ \cline{1-1}
\textbf{OpenAI (wo FT)} &  &  &  &  &  &  &\\ 
Text-Davinci-002&29.70&12.72&21.59&61.40&47.26&32.82&40.70\\ 
Text-Davinci-003&31.83&14.01&23.79&61.94&47.74&\textbf{41.10}&43.50\\ 
ChatGPT&35.98&14.09&23.46&62.26&48.43&39.53&43.68\\ 
GPT-4&38.63&14.11&23.95&63.81&\textbf{49.30}&39.48&44.54\\ 

\hline
\end{tabular}
\caption{Results of the summarization models on the \assessment{} division, test set 3. BART and LED models trained for full note generation perform suboptimally, likely as this content appearing later in a long sequence. Similar to test 1, the best models for \assessment{} are the BART Division models and the OpenAI Text-davinci-003 and GPT4 models.}
\label{tab:test3_assessment_and_plan}
\end{table}

\end{document}